\pdfoutput=1
\documentclass[dvipsnames,letterpaper, 10pt, conference]{ieeeconf}  

\usepackage{etoolbox}
\makeatletter
\patchcmd{\@makecaption}
  {\scshape}
  {}
  {}
  {}
\makeatother

\usepackage{placeins}

\IEEEoverridecommandlockouts

\usepackage[export]{adjustbox}
\usepackage{mathtools}

\usepackage[basicpaper,fig,lengthhacks]{tinyiu}

\microtypecontext{spacing=nonfrench}

\usepackage{xifthen}
\usepackage{tikz}
\usepackage{wrapfig}
\usepackage{minibox}

\DeclareCaptionStyle{figstyle}
[margin=0pt,justification=centering]
{format=hang,margin=15pt,
    font=scriptsize}
\definecolor{amethyst}{rgb}{0.6, 0.4, 0.8}
\definecolor{alizarin}{rgb}{0.82, 0.1, 0.26}
\definecolor{ashgrey}{rgb}{0.43, 0.5, 0.5}
\definecolor{yellow}{rgb}{1.0, 0.75, 0.0} %
\def\Cspace{\text{\emph{C-space}}\xspace}
\def\C{\ensuremath{C}\xspace}
\def\Cfree{\ensuremath{\C_\text{free}}\xspace}
\def\Cobs{\ensuremath{\C_\text{obs}}\xspace}
\def\Cobs{\ensuremath{\C_\text{obs}}\xspace}

\def\qinit{\ensuremath{q_\text{init}}\xspace}
\def\qtarget{\ensuremath{q_\text{target}}\xspace}

\makeatletter

\def\rrt{\textsc{rrt}}%
\newcommand*\RRT{%
    \@ifstar{%
        \rrt\textsuperscript{*}\xspace%
    }{%
        \rrt\xspace%
    }}

\def\rrdt{\textsc{rr}{\smaller d}\textsc{t}}%
\newcommand*\RRdT{%
    \@ifstar{%
        \rrdt\textsuperscript{*}\xspace%
    }{%
        \rrdt\xspace%
    }}

\def\prm{\textsc{prm}}%
\newcommand*\PRM{%
    \@ifstar{%
        \prm\textsuperscript{*}\xspace%
    }{%
        \prm\xspace%
    }}

\makeatother

\newcommand*\Von[1][d]{\boldsymbol{V}\hspace{-.3em}\boldsymbol{on}_{#1}}
\newcommand*\Prob{\mathbb{P}}

\newcommand*\FailedSamSet{\mathcal{X}}
\newcommand*\SampleDist{\mathcal{Q}}
\newcommand*\SampleDistMul{\hat{\SampleDist{}}} %
\newcommand*\SampleDistQ{f_q}
\newcommand*{\SampleDistDelta}{\Delta_{\SampleDistMul}}

\newcommand*\State{\Theta}
\newcommand*\state{\theta}
\newcommand*\Observable{Q}

\newcommand*\ExpectedGain{\mathcal{V}_\text{gain}}

\newcommand*\param{\theta}

\newcommand*{\norm}[1]{\lVert#1\rVert}

\newcommand*\vect[1]{\boldsymbol{#1}}
\newcommand*\given[1][]{\:#1\vert\:}

\makeatletter
\newcommand{\overrightsmallarrow}{\mathpalette{\overarrowsmall@\rightarrowfill@}}
\newcommand{\overarrowsmall@}[3]{%
  \vbox{%
    \ialign{%
      ##\crcr
      #1{\smaller@style{#2}}\crcr
      \noalign{\nointerlineskip}%
      $\m@th\hfil#2#3\hfil$\crcr
    }%
  }%
}
\def\smaller@style#1{%
  \ifx#1\displaystyle\scriptstyle\else
    \ifx#1\textstyle\scriptstyle\else
      \scriptscriptstyle
    \fi
  \fi
}
\makeatother

\newcommand*\Path[1]{\overrightsmallarrow{#1}}

\newcommand*\bLookOut[1]{%
    \ensuremath{\beta}\textsc{\footnotesize{-LookOut}}%
    \ifthenelse{\isempty{#1}}%
    {}{(#1)}%
}

\mathchardef\ordinarycolon\mathcode`\:
\mathcode`\:=\string"8000
\begingroup \catcode`\:=\active
  \gdef:{\mathrel{\mathop\ordinarycolon}}
\endgroup

\usepackage[boxed,ruled,vlined,linesnumbered]{algorithm2e}
\DontPrintSemicolon
\SetKwProg{Fn}{function}{}{}
\SetKwFunction{FnSampleFree}{SampleFree}
\SetKwFunction{FnRestartArm}{RestartArm}
\SetKwFunction{FnPickArm}{PickArm}
\SetKwFunction{FnRewire}{Rewire}
\SetKwFunction{FnNearest}{Nearest}
\SetKwFunction{FnRestartArm}{RestartArm}
\SetKwComment{Comment}{$\triangleright$\ }{}
\SetKwInput{KwInit}{Initialise}

\usepackage{cleveref}                                        
\crefname{assumption}{assumption}{assumptions}
\crefname{problem}{problem}{problems}
\crefname{algorithm}{Alg.}{Algs.}
\Crefname{algorithm}{Algorithm}{Algorithms}
\crefname{figure}{Fig.}{Figs.} 
\crefformat{equation}{(#2#1#3)}
\crefrangeformat{equation}{(#3#1#4) to~(#5#2#6)}
\crefmultiformat{equation}{(#2#1#3)}%
{ and~(#2#1#3)}{, (#2#1#3)}{ and~(#2#1#3)}

\usepackage[
    backend=biber
    ,giveninits=true                  
    ,url=false, isbn=false, doi=false 
    ,maxnames=99                       
    ,minnames=3                       
    ,sorting=none             
    ,date=year                
    ,labeldate=year
    ,style=ieee
]{biblatex}
\AtEveryBibitem{                      
    \iffieldequalstr{eprinttype}{jstor}
    {\clearfield{eprint}}
    {}
}

\AtEveryBibitem{%
    \clearfield{note}%
    \clearfield{keywords}%
    \ifentrytype{inproceedings}{%
        \clearfield{labelyear}%
        \clearfield{publisher}%
        \clearfield{volume}%
        \clearfield{urldate}%
        \clearfield{date}%
        \clearfield{pages}%
    }{%
    }%
}
\DeclareSourcemap{
    \maps{
        \map{
            \pertype{inproceedings}
            \step[fieldset=publisher, null]
            \step[fieldset=urldate, null]
            \step[fieldset=volume, null]
            \step[fieldset=pages, null]
        }
    }
}
\renewrobustcmd*{\bibinitdelim}{\,} 
\newcommand*{\autociteauthor}[1]{\citeauthor*{#1}~\autocite{#1}} 

\Crefname{figure}{Fig.}{Figs.}
\title{\LARGE \bf
Bayesian Local Sampling-based Planning
}

\author{Tin Lai$^{\dagger*}$, Philippe Morere$^{\dagger}$, Fabio Ramos$^{\dagger\mathsection}$ and Gilad Francis$^{\dagger}$
\thanks{
$^{*}$Correspondence to \texttt{\small tin.lai@sydney.edu.au}
}
\thanks{
$^{\dagger}$School of Computer Science, The University of Sydney, Australia.
}
\thanks{
$^{\mathsection}$NVIDIA, USA.
}%
}

\begin{document}

\maketitle
\thispagestyle{empty}
\pagestyle{empty}

\begin{abstract}
    Sampling-based planning is the predominant paradigm for motion planning in robotics.
    Most sampling-based planners use a global random sampling scheme to guarantee probabilistic completeness.
    However, most schemes are often inefficient as the samples drawn from the global proposal distribution, and do not exploit relevant local structures.
    Local sampling-based motion planners, on the other hand, take sequential decisions of random walks to samples valid trajectories in configuration space.
    However, current approaches do not adapt their strategies according to the success and failures of past samples.

    In this work, we introduce a local sampling-based motion planner with a Bayesian learning scheme for modelling an adaptive sampling proposal distribution.
    The proposal distribution is sequentially updated based on previous samples, consequently shaping it according to local obstacles and constraints in the configuration space.
    Thus, through learning from past observed outcomes, we maximise the likelihood of sampling in regions that have a higher probability to form trajectories within narrow passages.
    We provide the formulation of a sample-efficient distribution, along with theoretical foundation of sequentially updating this distribution.
    We demonstrate experimentally that by using a Bayesian proposal distribution, a solution is found faster, requiring fewer samples, and without any noticeable performance overhead.

\end{abstract}

\section{Introduction}

Motion planning is a critical aspect in accomplishing most robotic tasks, from vacuum cleaning to needle insertion.
It involves planning the trajectories of the actuated parts of the robot, under joints or motion constraints, while avoiding  collision with obstacles, transiting the system into a desired state.
In short, a motion planner produces a feasible trajectory that is safe, and possibly minimising some cost, such as distance or energy consumption.

The computational complexity of a motion planner depends on the dimensionality of \Cspace---the configuration space that encapsulates the set of all possible configurations---defined by the number of actuated joints.
While more joints offer greater flexibility in motion, the additional degrees-of-freedom require additional computing resources for planning.
Consequently, motion planning in a high-dimensional configuration space is an active field of research~\autocite{al-bluwi2012_MotiPlan}.
Indeed, the famous $A^*$ algorithm has a complexity $\mathcal{O}(b^d)$~\autocite{russell2016_ArtiInte} which is exponential in $d$, the dimensionality of the search space\footnote{$b$ is the branching factor, the average number of children at each node}.
Therefore, any attempts to search for a valid trajectory directly in the \Cspace is intractable.

Sampling-based planners (SBPs) are a class of motion planners that provide a robust approach to trajectory planning~\autocite{elbanhawi2014_SampRobo}.
The objective of such a planner is to avoid the explicit construction of \Cspace because of intractability~\autocite{lavalle2006_PlanAlgo}.
Instead, such a planner samples configurations randomly and builds a graph or tree-like structure that captures valid connections between different configurations.
This structure is then used to finds a valid trajectory to transits the system to the desired state.

\begin{figure}[t]
    \centering%
    \begin{subfigure}[t]{.5\linewidth}
        \centering
        \includegraphics[width=\linewidth]{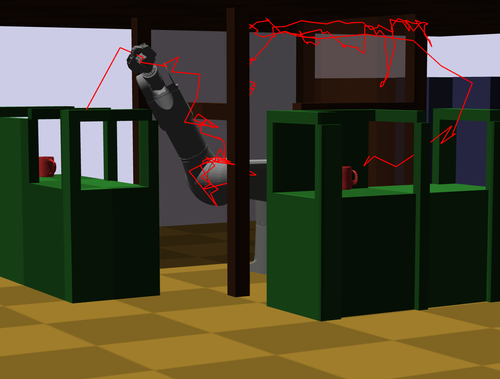}
    \end{subfigure}%
    \begin{subfigure}[t]{.5\linewidth}
        \centering
        \includegraphics[width=\linewidth]{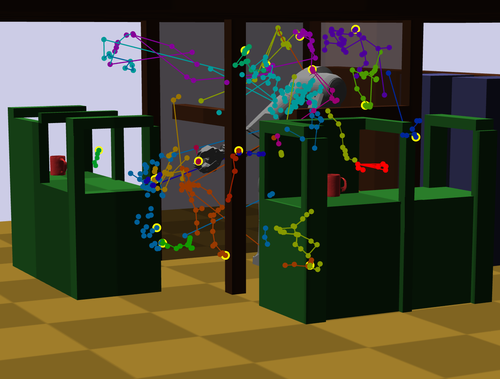}
    \end{subfigure}
    \caption{
    Experiment scenario with the TX90 manipulator.
    The robotic arm needs to transit its gripper, through the narrow openings of the cluttered beams and columns, to connect the two red cups without collision.
    Collisions refer to obstacles collision, self collision, and TX90's joints rotational limits.
    A possible solution trajectory (without nodes for clarity) is shown on the left.
    Each colour on the right illustrates a corresponding disjointed-tree in our algorithm, projected the configurations into the obstacle space.
    The overlaid nodes and edges illustrate the spatial location of the end effector (the PR2 gripper).
    \label{fig:robotic-arm-expr}
    }
\end{figure}

Most SBPs are \emph{probabilistically complete}---that is, the probability of failing to find a feasible solution, if one exists, converges to zero exponentially fast~\autocite{kavraki1996_AnalProb}.
While SBPs are probabilistically complete, their runtime is limited by the complexity of \Cspace, primarily due to narrow passages~\autocite{hsu1998_FindNarr} that limit the connectivity of the free space in the highly restricted regions.
The challenge in planning with narrow passages is that the probability of an event of successfully extending a connection from the initial configuration into a narrow passage is very low;
moreover, such an event needs to occur multiple times in nearby regions for the entire trajectory along the narrow passage to be built.
\Cref{fig:robotic-arm-expr,fig:robotic-arm} illustrate the manipulation of a robotic arm, with \Cref{fig:robotic-arm} highlighting how planning a trajectory in workspace translates into multiple narrow passages in configuration space.

{
\setlength{\intextsep}{0mm}
\begin{figure*}[tb]
    \centering
    \subcaptionbox{%
        A 3-dof robotic arm moving in workspace \label{fig:robotic-arm:obstacle-space}}%
    {%
        \begin{tikzpicture}
            \node (img) {\includegraphics[height=0.24\textheight]{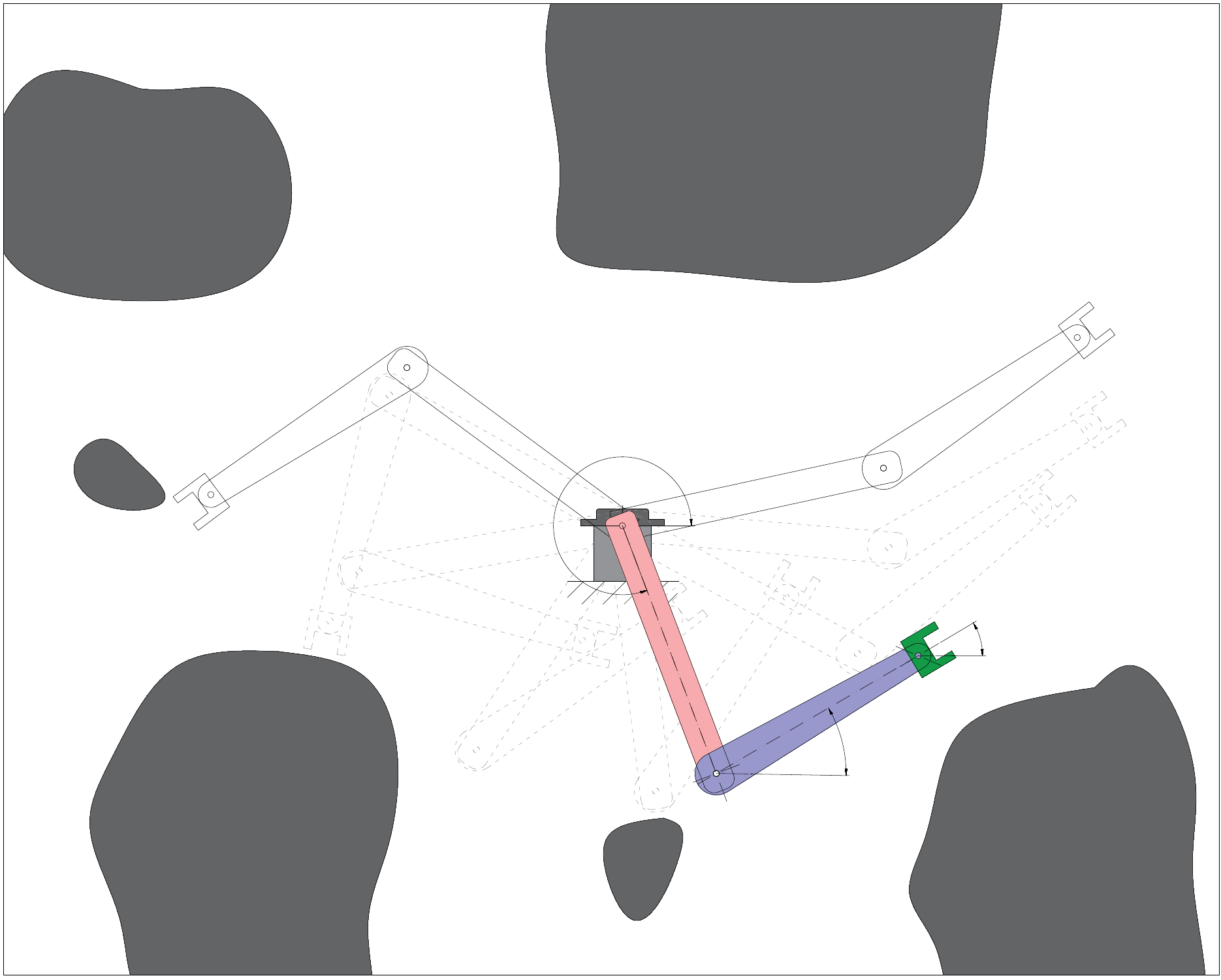}};

            \node[xshift=.2cm, yshift=.5cm] (t1) {$\phi_0$};
            \node[align=left,xshift=1.58cm, yshift=-1.45cm] (t2) {$\phi_1$};
            \node[align=left,right = .155cm of t2, yshift=.66cm] (t3) {$\phi_2$};

            \node at (.8, -2) (qinit) {\color{purple}\small$q_\text{init}$};
            \node at (-2, .48) (q1)    {\color{purple}\small$q_i$};
            \node at (2, .9) (q2)    {\color{purple}\small$q_j$};
        \end{tikzpicture}
    }
    \hspace{.1cm}
    \subcaptionbox{%
        Corresponding trajectories in \Cspace
        \label{fig:robotic-arm:configuration-space}}%
    {%
        \begin{tikzpicture}
            \node (img) {\includegraphics[height=0.24\textheight]{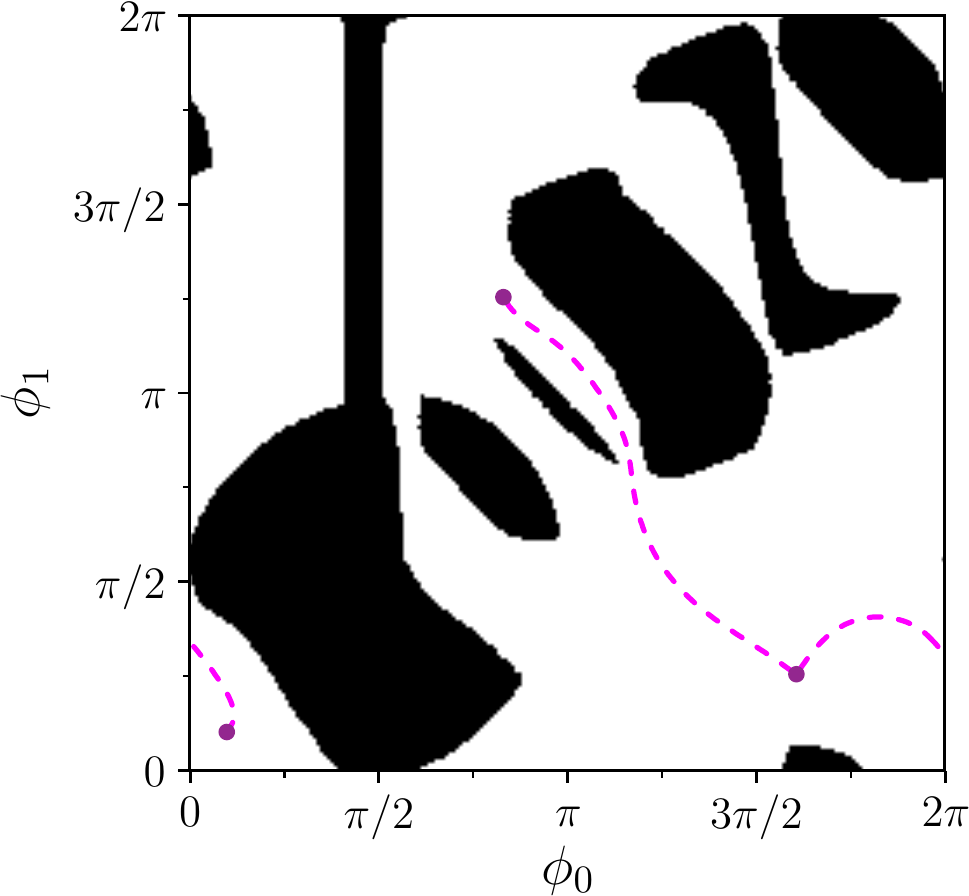}};
            \node at (1.7, -1.55) (qinit) {\color{purple}\small$q_\text{init}$};
            \node at (-1.4, -1.75) (q1)    {\color{purple}\small$q_j$};
            \node at (-.15, .9) (q2)    {\color{purple}\small$q_i$};
        \end{tikzpicture}%
    }
    \caption{
         Planning two trajectories (from $q_\text{init}$ to $\{q_i, q_j\}$) for a 3 degree of freedom (dof) robotic arm, with its 3 rotational joints $\{\phi_i\}^2_{i=0}$, in (\subref{fig:robotic-arm:obstacle-space}) workspace and (\subref{fig:robotic-arm:configuration-space}) \Cspace.
        While obstacles in the workspace view do not seem to obstruct the motion of the arm, \Cspace view demonstrates that planning over feasible configurations is very obstructed. Thus, increasing sampling efficiency around narrow passages will remove such bottlenecks during planning.
        (Note that we fixed $\phi_2 = \phi_1$ for 2D projection, and \Cspace view is shown solely for illustration purposes and is not available during planning).
        \label{fig:robotic-arm}
    }
\end{figure*}
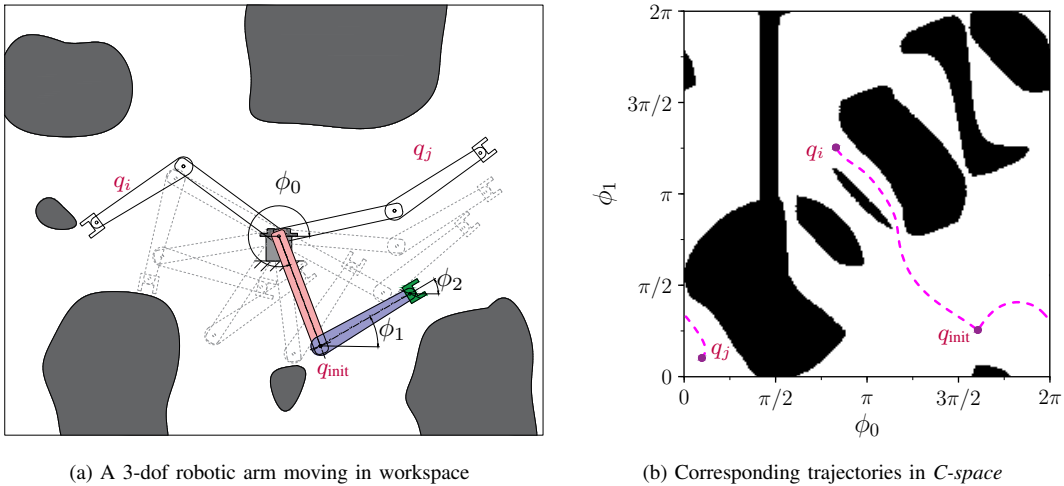
}

In this paper, we focus our attention on sampling efficiently in narrowed spaces.
In particular, our work sequentially updates its local proposal distribution of sampling points, such that it adaptively adjust its distribution based on \Cspace's local structures.
Our work utilises Bayesian update rule for sequentially adapting the proposal distribution according to observing past sampled results.
Our contributions are as follows.
First, we formulate the sequential sampling problem as a Markov process and demonstrate the necessity of a sequentially updated proposal distribution by incorporating past results.
Second, we provide a sampling framework that utilises kernel functions for updating the proposal distribution.
The Bayesian updates is incremental which is suitable in our incremental planning setting.
Third, we provide an efficient approach for performing sequential updates on the proposal distribution, and drawing new random samples from it.
Lastly, we provide empirical results that reinforce the claim that better sample-efficiency in motion planning can be achieved by utilising Bayesian proposal distribution.

\section{Background}

Notable SBPs include Probabilistic Roadmap (\PRM), proposed by \autociteauthor{kavraki1996_ProbRoad} which provides solid theoretical foundations on probabilistic completeness to the class of SBPs.
\autociteauthor{lavalle1998_RapiRand} proposed another class of SBPs with Rapidly-exploring Random Tree (\RRT), which is an anytime algorithm for single query planning.
\PRM* \autocite{karaman2010_IncrSamp} and \RRT* \autocite{karaman2011_SampAlgo}, as the star variants of the previous two algorithms, denote an \emph{asymptotic optimality} guarantee \autocite{elbanhawi2014_SampRobo}, which means trajectories found by these planners converge to the optimal solution almost surely.

The problem of sampling within narrow passages is a widely recognised problem in robotics~\autocite{hsu1998_FindNarr, elbanhawi2014_SampRobo,sun2005_NarrPass,wang2010_TripRRTs}, and numerous strategies were developed to address it, including using bridge test to locate narrow passages~\autocite{wilmarth1999_MAPRProb,hsu2003_BridTest}, adaptively bias toward regions with limited visibility~\autocite{yershova2005_DynaRRTs}, or utilise optimisation technique to have higher likelihood to generate samples close to obstacles' boundary~\autocite{zhang2008_EffiRetr,lee2012_SRRRSele}.
However, utilising heuristic to discover narrow passages is not a trivial task as obstacles have no explicit representation in \Cspace.

Several authors proposed the importance of a balance between exploration and exploitation during planning~\autocite{rickert2008_BalaExpl,lai2018_BalaGlob}, borrowing the idea from game theory literature.
The majority of research is based on using heuristic biasing to guide the search, with user-defined probability for random exploration to retain completeness.
For example, \autocite{ichter2018_LearSamp} proposed the use of a conditional variational autoencoder to learn the manifolds of the regions for generating samples that are more likely to be successful.
However, such a method requires a strategy to maintain random exploration to ensure completeness.
The search for connections can also be split into a 2-step procedure, first with initial workspace exploration, followed by an exploitative stage utilising previous knowledge~\autocite{urmson2003_ApprHeur}.
There exists technique that characterise and partition \Cspace in regions that deems to be favourable to certain type of planners~\autocite{morales2004_MachLear}.
Such a method recognising local-structures, but is dependent on the ability of correctly classifying the local regions and assigning the more favourable planner.
Methods that utilise a different sampling sequence, for example the low-discrepancy methods which provide several advantage by the use of quasi-random sequences~\autocite{lindemann2003_IncrLowd}, can be used in complement with our planner for the theoretical guarantee that deterministic sequence provides.

Strategies utilising multiple exploring trees were also presented, such as growing bidirectional trees~\autocite{kuffner2000_RRTcEffi},
heuristically selecting a tree from a pool of local trees~\autocite{strandberg2004_AugmRRTp},
and using a learning technique to initialise trees in narrow passages probabilistically~\autocite{wang2018_LearMult}.
Deciding which trees to extend the connection from is also one of the dilemmas in utilising multi trees, where several authors propose to structure the selection as a multi-armed bandit problem for maximising gains.
\autociteauthor{lai2018_BalaGlob} employed multiple local sampling-based planners to perform sequential Markov chain Monte Carlo (MCMC) random walks to recover connectivity between states in free space. Scheduling sampling between the local planners is then formulated as a multi-armed bandit (MAB) problem. While MAB balances the exploration-exploitation that ensures efficient planning, the proposal distribution used by the MCMC random walk does not adapt according to the success and failures of past samples.

In this work, we tackle the shortcoming of the local sampling-based planning problem, by employing a sequential Bayesian update of the proposal distribution, taking full advantage of the sampled data.
Unlike previous work where the proposal distribution of each local planner uses a stationary distribution, we model the sequential random walk as a Markovian process where its proposal distribution is sequentially updated following collision of the random walker with obstacles. Consequently, the adaptive proposal distribution density in the vicinity of obstacles drops, leading to a higher likelihood of drawing the next sample in free space.
We formulate the local planning problem in the following section.

\section{Local Sampling-based Planning Problem}


The objective of a motion planner is to construct a trajectory from an initial configuration \qinit to a target configuration \qtarget, where $q\in \C \subseteq \mathbb{R}^d$ denotes a state in \Cspace with $d \ge 2$ denoting \Cspace dimensionality.
We use $\Cobs \subseteq \C$ to denote invalid states, and the set of all valid states is defined as the closure set of $\Cfree := \text{cl}(\C\setminus\Cobs)$.
In this work, we focus our attention on addressing the problem of planning in narrow spaces with with Bayesian sampling-based local planning.
%
%
%

\subsection{Exploring \Cspace with local planners}

\newcommand{\K}{\mathcal{K}}

\RRdT*~\autocite{lai2018_BalaGlob} formulates the sampling-based planning problem as a balance between the global exploration (global unseen spaces) and local-connectivity exploitation (local free spaces connectivity).
The goal is to build a graph $G=(V,E)$ that connects the initial state to the target state.
The balance of the two objectives is formulated as a multi-armed bandit (MAB) problem, where $k \in \K$ refers to the $k$ local planners used to exploit local connectivity.
Each local planner performs an MCMC random walk in \Cfree, such that the Markov Chains created by the random walks map out the connectivity of \Cfree.
With the connections formed by the Markov Chains---in the forms of nodes and edges---the planner resolves a valid and safe path from $q_\text{init}$ to $q_\text{target}$ as $N \to \infty$, where $N$ denotes the number of sampled configurations by the SBP.

Each local planner (chain)  is initialised at some $q_k\in\Cfree \, \forall k \in \K$. The chain explores $\Cfree$ emulating a random walk with drift, where the direction of the random walk is sampled from a proposal distribution $h$. The chain then tries to expand by taking an $\epsilon$-step in a direction sampled from the proposal distribution. If the connection is not in \Cfree, due to a self or obstacle collision, the chain's expansion is rejected, and the process repeats with a new direction sampled from $h$. If the proposed step is valid, the chain expands to the new configuration whilst updating $G$ with the new node and edge.

This formulation of local sampling-based planning is proven to be probabilistically complete and asymptotic optimal as $N \to \infty$.
However, the local proposal distribution introduced by \autocite{lai2018_BalaGlob} is only conditioned on previous successful samples from $h$, without considering the rejected samples.
Hence, re-sampling following a rejected sample was drawn from the same fixed proposal distribution. The re-sampling process continued until a successful sample was drawn or until the MAB scheduler re-initialise the local planner in some other $q \in \Cfree$.
In this work, We propose a Bayesian learning framework for the proposal distribution, which incorporates invaluable information in both successful and rejected samples. The proposal distribution $\SampleDist(\cdot \given  \FailedSamSet)$ is sequentially updated following failed samples $\FailedSamSet$, thus adapting to the shape of local constraints. As a result,  $\SampleDist$ can propose more promising directions, with an increased probability of success.

\section{Sequential Bayesian Updates on Proposal Distribution}

In this section, we formulate local sampling-based planning as a Markov process, and propose a sequential Bayesian updating for the proposal distribution.

\subsection{Motivation}\label{sec:motivation}

\begin{figure}[tb]
            \centering%
            \begin{tikzpicture}[every node/.style={inner sep=0,outer sep=0}] 
                \node (img) {\includegraphics[width=0.5\linewidth]{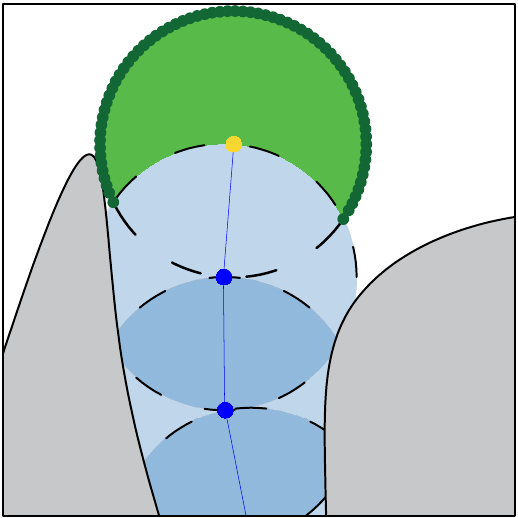}};
                \node at (1.3, -1.2) {\footnotesize$\Cobs$};
                \node at (-1.5, -.8) {\footnotesize$\Cobs$};
            \end{tikzpicture}%
            \begin{tikzpicture}[every node/.style={inner sep=0,outer sep=0}] 
                \node (img) {\includegraphics[width=0.5\linewidth]{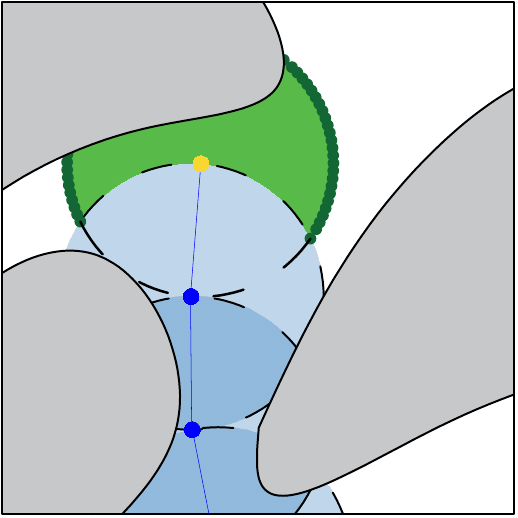}};
                \node at (-1.2, 1.5) {\footnotesize$\Cobs$};
                \node at (1.2, -.5) {\footnotesize$\Cobs$};
                \node at (-1.2, -1.0) {\footnotesize$\Cobs$};
            \end{tikzpicture}
            \caption{
            Local planning in \Cspace.
            \textcolor{BurntOrange}{Yellow node} represents current state $\State_n$ of the local planner,
            \textcolor{blue}{blue node} represents previous states $\{ \State_i \}_{1 \le i < n}$,
            with the \textcolor{blue}{edges} representing the chain constructed by the planner.
            The \textcolor{TealBlue!40!blue}{cyan regions} represent the $\epsilon$-ball volume of previously sampled states,
            and \textcolor{Green!80!black}{green regions} being the transitable, unexplored, states in $\Cfree$.
            (left) Local planning in the absence of obstacles.
            (right) Cluttered environment where the local planning should sequentially adapt its proposal distribution according to the surrounding obstacles.
            \label{fig:chains-transition}
            }
        \end{figure}
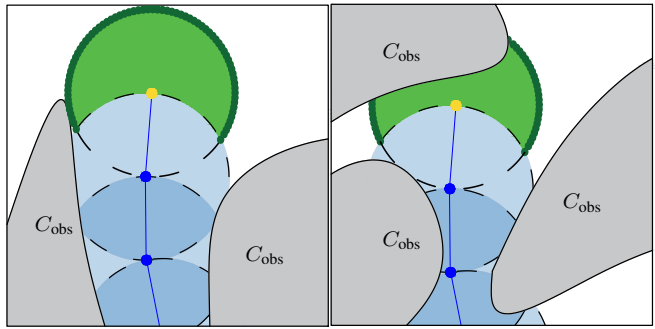

The objective of local planners in our setting is to maximise coverage of unexplored space,
where we define the volume of unexplored free spaces gained by sampling $q$ as
\begin{equation}
    \ExpectedGain(q) = \epsilon(q) \cap \Cfree \setminus \bigcup_{v \in V} \epsilon(v),
\end{equation}
where $\epsilon(v)$ denotes the $d$-dimensional unit sphere with a radius of $\epsilon$ centred at $q$.
The $\ExpectedGain(q)$ denotes the volume of unexplored free spaces that we gain by sampling $q$.
This process is illustrated in~\cref{fig:chains-transition}, where the green shade identifies the gain volume $\ExpectedGain(q)$ after sampling the yellow configuration point.

Assuming that we have complete knowledge of the transitional dynamics of \Cspace.
Then, we can formulate the optimal tree expansion as selecting an optimal transition function $a^*$ that maximise the coverage in the exploration sequence with
\begin{multline}
    a^* = \argmax_{a \in \mathcal{A}}
     [
        \ExpectedGain(q_{t}) + \gamma \ExpectedGain(q_{t+1}) + \\ \gamma^2 \ExpectedGain(q_{t+2}) + \cdots
        ]
    \text{~~for~~} \gamma\in[0,1),
    \label{eq:max-exploration_new}
\end{multline}
where the transition function $a\in\mathcal{A}$ maps a given configuration state $q_{n}$ to $q_{n+1}$, and $\gamma$ is a discount factor for future gains.

        \setlength\intextsep{0pt}
        \captionsetup{belowskip=0pt}
        \setlength{\belowcaptionskip}{-10pt}
        In practice, Eq.~\cref{eq:max-exploration_new} cannot be directly solved as there is no closed-form transition dynamic in \Cspace, meaning the transition function is unknown.
        Instead, we replace the direct optimisation problem with sampling, where the new samples are drawn from an adaptive proposal distribution $q_\text{new} \sim \SampleDistQ(q \given \param, \mathcal{D})$.
        The distribution $\SampleDistQ$ captures the probability of drawing a successful sample.
        The parameters $\param$ of the proposal distribution are updated sequentially according to the success and failures of previous samples, and $\mathcal{D}$ refers to the observed sample results.
        Consequently, the direct objective function defined in ~\cref{eq:max-exploration_new} can be replaced with an optimisation over the expected value as defined in
\begin{multline}
    q^*_\text{new} = \argmax_{q_t \sim f_q} ~
    \mathbb{E} [
        \ExpectedGain(q_{t}) + \gamma \ExpectedGain(q_{t+1}) + \\
         \gamma^2 \ExpectedGain(q_{t+2}) + \cdots
        ].
    \label{eq:max-exploration}
\end{multline}
The expectation taken in~\cref{eq:max-exploration} is with respect to the stochasticity from transiting states by sampling $f_q$, as we do not have direct access to the transition function in~\cref{eq:max-exploration_new}.

        In the next section, we formulate the local planning problem as a Markov process with unobservable state, and subsequently model a proposal distribution that encapsulates the Markovian property by proposing tree extensions that incorporate previous successful and failed samples.

\subsection{Local Planning as a Markovian process}\label{sec:markovian-process}

We model the sampling procedure as a Markovian process.
A state $\State_n$ of a local planner refers to its spatial location and the properties of \Cspace that are nearby at step $n$.
It determines how likely it is to extends a connection in $\State_n$ to nearby states, and can be regarded as a tuple $(q_n,(V,E),\Cfree,\Cobs)$ of its current location, observed information, and the environment itself.
We cannot directly observe the state $\State_n$ of the local planner---where it is dependent on where in \Cspace it is at---but we can observe the outcome by sampling directions $x_\text{new}$ from local planner's current location $q_n$ to extends towards a nearby configuration $q_\text{new}$.
We call that an observation outcome is a successful extension if $\Path{q_n q_\text{new}}\in\Cfree$, and $\Path{q_n q_\text{new}}\notin\Cfree$ as a failed extension, where the notation $\Path{q_i q_j}$ denote the connection between the configurations $q_i$ and $q_j$.

In the formulation of \RRdT* local planning, the transition probability of a local planner depends solely on its current location and its last successful sampled direction, but not on its history.
That is, the transitional probability of local planners obey the Markovian property, such that its conditional probability distribution of $\State_k$ depends solely on $\State_{k-1}$.
Let $\State_i = \state_i$ denotes the event that $\State_i$ is at $\state_i$ at step $i$.
The state $\State_n$ and sampled point $\Observable_n = q_n$ are discrete-time stochastic processes, where the pair $(\State_n,\Observable_n)$ can be written as
\begin{multline}
    \Prob(\Path{q_n q_\text{new}} \in \Cfree \given \State_1 = \state_1, \ldots , \State_n = \state_n) \\ = \Prob(\Path{q_n q_\text{new}} \in \Cfree \given \State_n = \state_n)
\end{multline}
for every $n \ge 1$ and every $\state_1,\ldots,\state_n$;
with $q_\text{new}$ being the new proposed configuration extension by extending $q_n$ an $\epsilon$ amount towards some direction $x_\text{new}$.

\Cref{fig:chains-transition} demonstrates two local planners having the same set of transitional observation $\{x_i\}_{i=1}^{n}$ on which directions $x_i$ can successfully extend the tree (Notice that the angles of each chains are same for both planners).
However, the two planners consist of different sequence of hidden states $\{ \State_i \}_{i=1}^{n}$, where it is conditional on previous successful direction $x_{i-1}$ and can exhibit substantially different outcomes (the left planner in \cref{fig:chains-transition} has more \Cfree to explores while the other has a lower probability of successful tree extension).
The question remains how to find the right direction $x_\text{new}$ to extend our $q_n$ towards new configurations, such that it will have a higher probability for a successful extension.

In the following, we present a method that uses Bayesian update rule for modelling the proposal distribution for proposing new configurations.



\begin{algorithm}[tb]
    \caption{Bayesian local sampling} \label{alg:bayesian-local-sampling}
    \KwIn{$q_{init},q_{goal},N,\SampleDistMul$}
    Initialise $k$ arms (local samplers) into $\mathcal{K}$; $n \gets 1$ \;
    \While(\Comment*[f]{max number of nodes}){$n \le N$}{
        Add $k\in\mathcal{K}$ with low probability to restart-queue \;
        Restarts $k$ in restart-queue if it has any \;
        \uIf{local sampler $k\in\mathcal{K}$ restarted}{ \label{alg:rrdt:restart-arm}
            $n \gets n + 1$\;
        }
        \Else{
            $k \gets $ pick local sampler from $\mathcal{K}$ \Comment*[r]{MAB}
            $q_\text{current} \gets$ current position of $k$ \; \label{alg:rrdt:baye-sam-start}
            $\FailedSamSet \gets$ previous failed directions of $k$ in $q_\text{current}$ \;
            \uIf{$k$ has no previous history}{
                $x_i \sim \mathcal{U}(-\pi, \pi)^d$ \;
            }
            \Else{
                $x_{i-1} \gets $ last successful direction of $k$ \;
                $x_i \sim \SampleDistMul(x \given x_{i-1}, \FailedSamSet)$ \Comment*[r]{Eq. \ref{eq:full-proposing-baye-samp-dist}} \label{alg:rrdt:seq-update}
            }
            $q_\text{new} \gets q_\text{current} + \epsilon_n \cdot x_i$ \; \label{alg:rrdt:step-towards}
            \uIf{path from $q_\text{current}$ to $q_\text{new}$ is free}{
                Update $k$'s position to $q_{new}$\; \label{alg:rrdt:update-arm-pos}
                \If{can connect to other trees within $\epsilon_n$}{
                    Connect $k$'s tree to those trees \;
                    Add $k$ to restart-queue \;
                }
                $n \gets n + 1$\;
            }
            \Else{
                $\FailedSamSet \gets \FailedSamSet \cup q_i$ \;\label{alg:rrdt:baye-sam-end}
            }
            Updates $k$'s probability\;\label{alg:rrdt:a-arm-pos}
        }
    }
\end{algorithm}

\subsection{Modelling the Proposal Distribution}

First, let us introduce the underlying proposal distribution that we are modelling.
Instead of directly sampling configuration $q_\text{new} \sim \SampleDistQ(q \given \cdot)$ for local planning, which loses spatial information, we will sample a unit directional vector $x_\text{new} \sim \SampleDist(x \given \cdot)$ from current location $q_n$, and subsequently construct $q_\text{new} = q_n + \epsilon \cdot x_\text{new}$ as our new proposing configuration point.
The distribution $\mathcal{Q}$ is a a parametric function of $f_q$, that implicitly constrain the $q_\text{new}$ to be spatially close to the local planner to eliminate the discussed issues, while allowing the proposal distribution $\SampleDist$ to be flexible on its representation.
In general, $\SampleDist$ can be taken from any form of distribution that exploits the local structure of the narrow passage, given past events.

We formulate the distribution as one that incorporates information obtained from past failed samples and sequentially updates its posterior.
In our approach, $\SampleDist$ is modelled with directional distribution with Bayesian sequential updates.
Each local planner will sample next extension direction $x_i,j$ from a directional proposal distribution $\SampleDist_i(x \given x_{i-1}, \FailedSamSet_j)$ for the $j$\textsuperscript{th} sample in state $\State_i$, where $\FailedSamSet_j$ is the set of all failed directions at state $\State_i$.
When a local planner first visits the state $\State_i$, its proposal distribution will be set to its prior distribution $f_\text{prior}$, along with $\FailedSamSet_1 := \emptyset$.
Each time a local planner sampled a new direction $x_k$ and failed to extends its tree, we set $\FailedSamSet_j \gets \FailedSamSet_{j-1} \cup \; x_k$;
whereas when a local planner successfully extend its tree, it will transits from state $\State_i$ to $\State_{i+1}$.
The sampling procedures will be discussed in details in the following section.

\subsection{Sequential Bayesian Update}

The sequential nature of the local sampling-based planning problem comes naturally once it was identified as a Markovian process in~\cref{sec:markovian-process}.
Since the arrival to state $\State_i$ depends from its previous state $\State_{i-1}$, we can model the sampling procedure as a sequence of directional sampling where the direction $x_{i}$ depends on its previous successful direction $x_{i-1}$.
The sequential Bayesian updates come into play when we first initialise our proposal distribution $\SampleDist_1$ with our prior $f_\text{prior}$, and as the sampled result arrives sequentially, we update $\SampleDist_i$ with the result of $x_i$, to improve our likelihood to sample in more promising directions.

The current formulation allows any arbitrary directional distribution in the $N$-dimensional sphere to be used as the prior belief, for example, it could be as simple as a uniform distribution.
In our settings, we use the von Mises-Fisher distribution $\Von(\cdot)$~\autocite{fisher1995_StatAnal} as our prior distribution $f_\text{prior}$.
The $\Von$ distribution is a continuous probability distribution that is a close approximation to a wrapped directional Gaussian distribution.
The benefit of $\Von$ is twofold:
(i) it is the most mathematically tractable circular distributions which allows simpler statistical analysis, and
(ii) it is analogous to a ``circular normal distribution'' as it share the statistical benefits of being the limiting case for the sum of large number of angular deviations.
The $\Von$ distribution acts as our inductive bias to be explorative in our sampling.

Our prior $f_\text{prior}$ is coupled with a kernel $k(\cdot)$ that incorporates the failure information that we obtained from sample at $i-1$.
Moreover,
we initialise our proposal distribution such that $\SampleDist_1(x) = \Von(\vect{x} \given \vect{\mu}, \kappa)$.

The von Mises-Fisher probability distribution in $\mathbb{R}^d$ with $d \ge 2$ is given by
\begin{equation}
    \Von(\vect{x} \given \vect{\mu}, \kappa) = C_d(\kappa) e^{\kappa \vect{\mu}^T \vect{x}}
\end{equation}
where $\vect{\mu}$ is the unit vector of mean direction, $\kappa$ is the concentration parameter, $C_d(\kappa)$ is the normalising constant given by
\begin{equation}
    C_d(\kappa) = \frac{
    \kappa^{d/2-1}
    }{
    (2\pi)^{d/2}I_{d/2-1}(\kappa)
    },
\end{equation}
and $I_v(\cdot)$ is the modified Bessel function of order $v$.

\begin{figure}[b]
    \centering%
    \begin{subfigure}[t]{.41\linewidth}
        \centering
        \includegraphics[width=\linewidth]{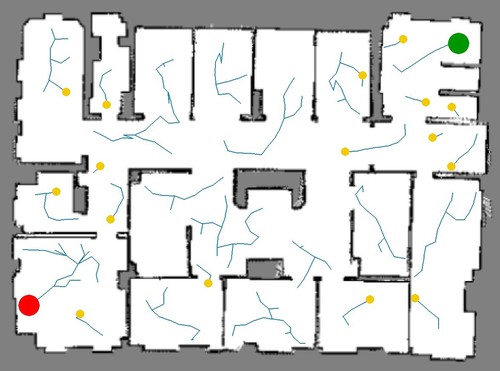}
    \end{subfigure}%
    \hspace*{.25em}
    \begin{subfigure}[t]{.3\linewidth}
        \centering
        \includegraphics[width=\linewidth]{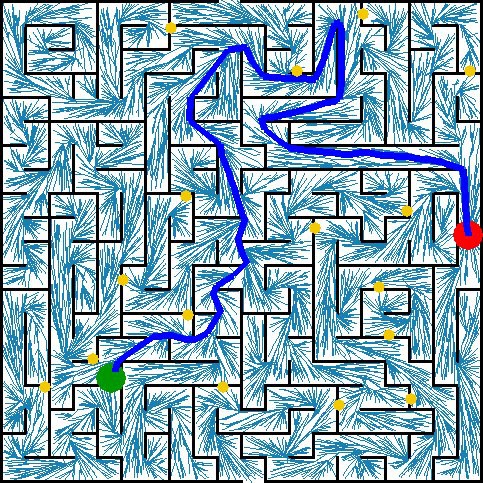}
    \end{subfigure}%
    \hspace*{.25em}
    \begin{subfigure}[t]{.145\linewidth}
        \centering
        \includegraphics[width=\linewidth]{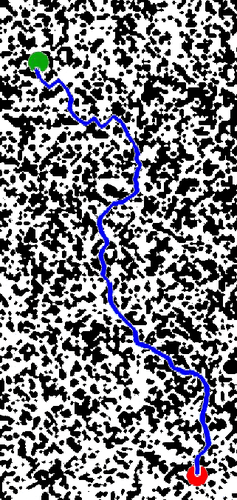}
    \end{subfigure}%
    \caption{
    Experimental scenarios. (Left) \emph{Room}, (Middle) \emph{Maze} and (Right) \emph{Clutter}; each with increasing complexity.
    \label{fig:all-expriment-envs}
    }
\end{figure}

We can draw $\vect{x} \sim \Von(\vect{x} \given \vect{\mu}, \kappa)$ to get a unit vector where $\norm{\vect{x}} = 1$.
In particular, $\Von(\vect{x} \given \vect{\mu}, \kappa)$ reduces to a uniform density when $\kappa = 0$, and $\Von(\vect{x} \given \vect{\mu}, \kappa)$ tends to a point density when $\kappa \to \infty$.
Therefore, the choice of those parameters characterises how our proposal distribution $\SampleDist$ behaves for the samples drawn in the early stage.

We use $\SampleDist$ as the proposal distribution for our MCMC random walker in the \RRdT* local planning, where we update our proposal distribution depending on whether our previous sample is successful or not.
Here, we define a successful sample as the local planner being able to extend our Markov Chain connections an $\epsilon$ distance from the current configuration point.
The proposal distribution could be from a wide variety of distribution that exploits the local structure of a narrow passage.
In our formulation, the von Mises-Fisher distribution acts as our inductive bias of concentrating our sampling distribution towards a direction where we were successful before.
We define $x_i$ to be the successful direction when transiting from state $\State_{i-1}$.

At state $\State_1$, a local planner is first initialised without prior knowledge (no previous history).
Therefore, we sample a uniform random direction $\tilde{x} \sim \mathcal{U}(-\pi, \pi)^d$ and we define $x_0 := \tilde{x}$.
Then, our sampling scheme for local planner follows
\begin{equation}
    x_{i,j} \sim \SampleDist_i(x \given x_i, \FailedSamSet_j)
    ~~ \forall\, i,j \ge 1, \label{eq:sampling-scheme}
\end{equation}
lisand when a sampled direction is successful we transits the local planner to $\State_i$.

At each iteration, a local planner draws $x_{i,j}$ and attempts to extends its tree towards $x_{i,j}$.
If it is unsuccessful, the local planner remains at state $\State_{i-1}$, and another sample is drawn at random from the updated proposal distribution $x_{i,j+1} \sim \SampleDist_i(x \given x_{i-1}, \FailedSamSet_{j+1})$.
Whereas if local planner successfully extends its tree towards $x_{i,j}$, we say that $x_i := x_{i,j}$, and local planner will transits to state $\State_i$ and proceed to draw sample from $x_{i+1,1} \sim \SampleDist_{i+1}(x \given x_{i}, \FailedSamSet_{1})$ in the next iteration;
Therefore, the updates follow the usual expression for a Bayesian updating scheme where $f_\text{posterior} \propto f_\text{prior} \cdot f_\text{likelihood}$ and subsequently using current posterior as our next prior.

\begin{figure}[tb]
    \centering
    \begin{subfigure}{\linewidth}
        \includegraphics[width=\linewidth]{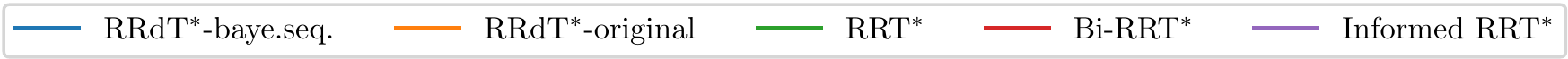}
    \end{subfigure}
    \begin{tikzpicture}

\node (img) {
    \begin{subfigure}[t]{.225\linewidth}
        \centering
        \includegraphics[width=\linewidth,trim={2mm 6.5mm 2mm 2mm},clip]
        {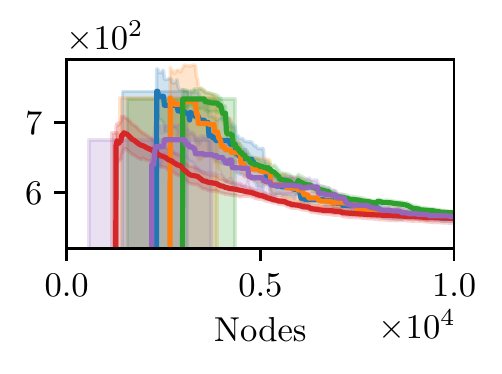}
        \includegraphics[width=\linewidth,trim={2mm 6.5mm 2mm 2mm},clip]
        {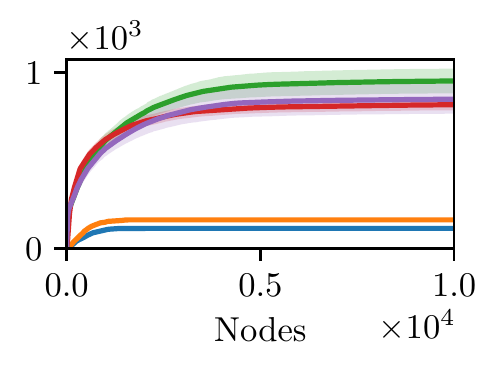}
        \includegraphics[width=\linewidth,trim={2mm 2mm 2mm 2mm},clip]
        {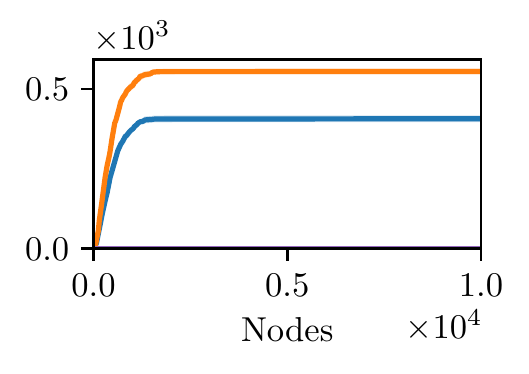}
        \caption{Room\label{fig:experiments:room}}
    \end{subfigure}%
    \begin{subfigure}[t]{.225\linewidth}
        \centering
        \includegraphics[width=\linewidth,trim={2mm 6.5mm 2mm 2mm},clip]
        {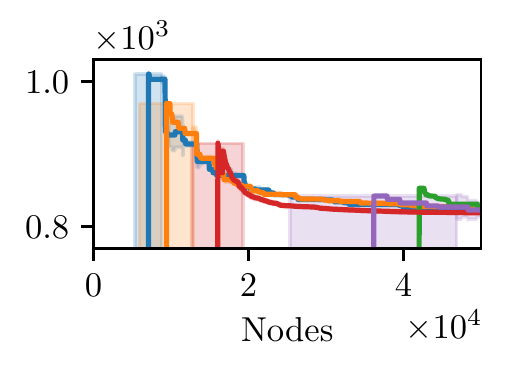}
        \includegraphics[width=\linewidth,trim={2mm 6.5mm 2mm 2mm},clip]
        {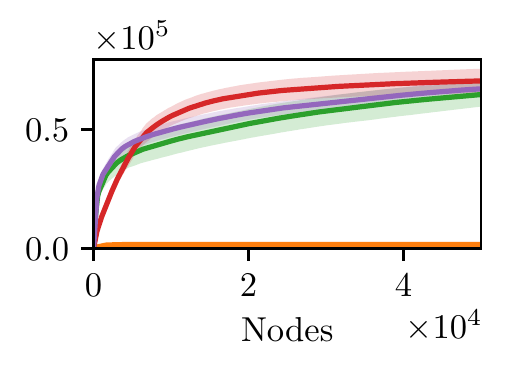}
        \includegraphics[width=\linewidth,trim={2mm 2mm 2mm 2mm},clip]
        {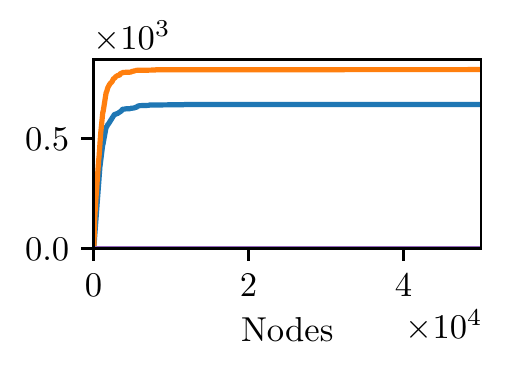}
        \caption{Maze\label{fig:experiments:maze}}
    \end{subfigure}%
    \begin{subfigure}[t]{.225\linewidth}
        \centering
        \includegraphics[width=\linewidth,trim={2mm 6.5mm 2mm 2mm},clip]
        {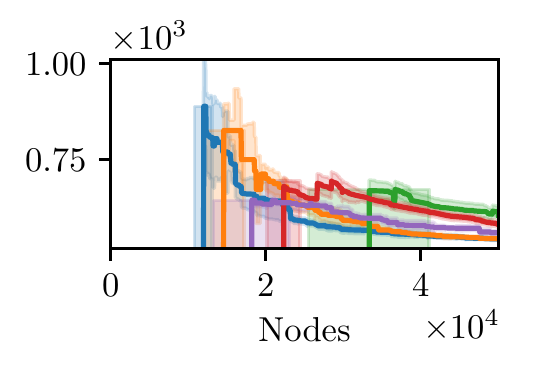}
        \includegraphics[width=\linewidth,trim={2mm 6.5mm 2mm 2mm},clip]
        {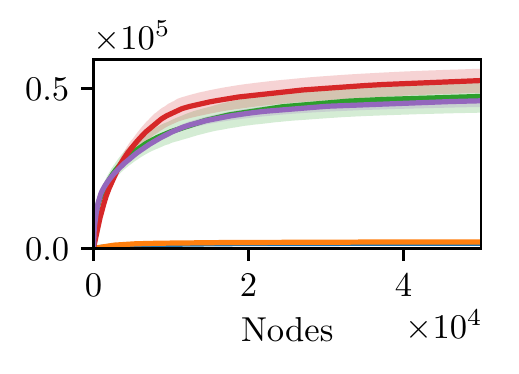}
        \includegraphics[width=\linewidth,trim={2mm 2mm 2mm 2mm},clip]
        {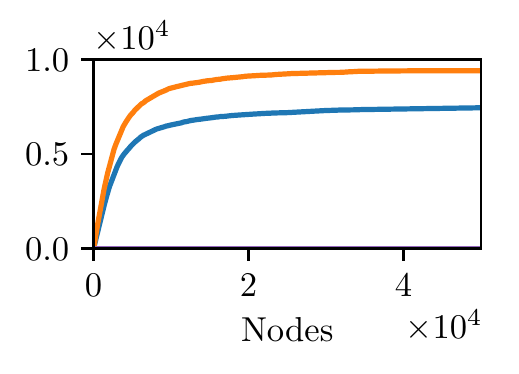}
        \caption{Clutter\label{fig:experiments:clutter}}
    \end{subfigure}%
    \begin{subfigure}[t]{.225\linewidth}
        \centering
        \includegraphics[width=\linewidth,trim={2mm 6.5mm 2mm 2mm},clip]
        {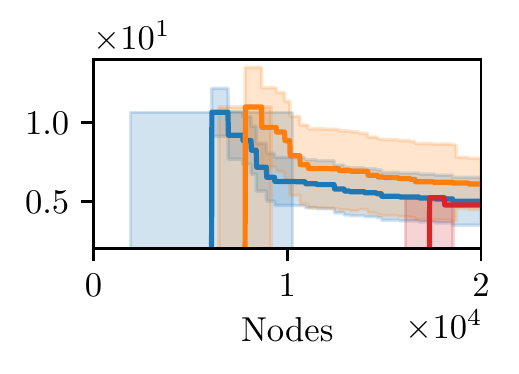}
        \includegraphics[width=\linewidth,trim={2mm 6.5mm 2mm 2mm},clip]
        {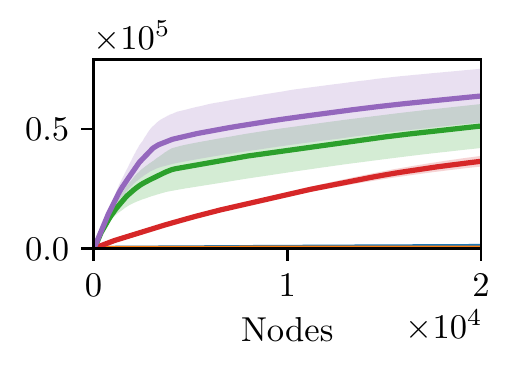}
        \includegraphics[width=\linewidth,trim={-.73mm 2mm 2mm 2mm},clip]
        {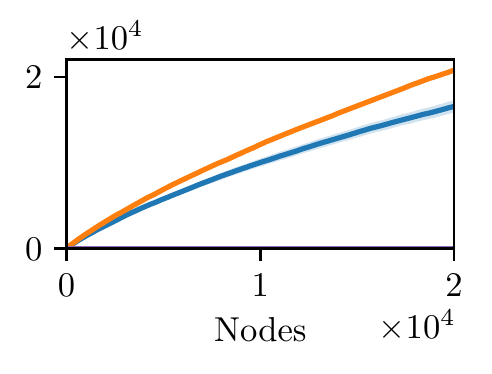}
        \caption{Manipulator\label{fig:experiments:manipulator}}
    \end{subfigure}%
};
\node [rotate=90] at (-4.1,1.4){
    \centering\footnotesize
    Cost
};
\node [rotate=90] at (-4.1,.2){
    \centering\footnotesize
    i.conn.
};
\node [rotate=90] at (-4.1,-1.2){
    \centering\footnotesize
    i.local.sam
};

\end{tikzpicture}%
    \caption{\footnotesize
        Comparison of \emph{cost} (top), \emph{invalid connections} (middle), \emph{invalid local sampling} (bottom) as a function of nodes in various planning scenarios, shaded region indicates the standard deviation.
        \label{fig:experiments}
    }
\end{figure}

\subsection{Proposal Distribution Update}

In the following, we will discuss how to utilise a periodic kernel $k_(\cdot)$ to sequentially incorporate past sampled results into $\SampleDist$.
The likelihood function is constructed to encapsulate the idea of having a decreasing nature to sampling again in previously failed directions.
We formulate the likelihood function as $f_\text{likelihood}(x \given x') \propto \big(1 - k(x, x')\big)$.
We can rewrite the posterior of the proposal distribution as
\begin{align}
    \SampleDist_i(x \given x_{i-1},\FailedSamSet_j)
     & = \frac{ \SampleDist_i(x \given x_{i-1},\FailedSamSet_{j-1})
    \left(1 - k(x,x'_{j-1}) \right)}{\alpha_{j}} ,
    \label{eq:recursive-def-sample-dist}
\end{align}
where $x'_{j-1} \in \FailedSamSet_{j-1} = \set{x'_{1}, \ldots x'_{j-2}}$ for $j > 1$
and $\alpha_{j}$ is the normalising factor.
Note that $\SampleDist_{i}(x \given x_{i-1}, \FailedSamSet_{1})$ reduces to $f_\text{prior}(x \given x_{i-1})$ as $\FailedSamSet_{1} = \emptyset$.

{
\addtolength{\dblfloatsep}{-50mm}
\addtolength{\textfloatsep}{-3mm}

\captionsetup{belowskip=-10pt}

\begin{figure*}[!bth]
    \centering
    \begin{subfigure}[t]{.12\textwidth}
        \noindent\makebox[\linewidth][c]{%
            \raisebox{-70pt}{\rotatebox[origin=c]{45}{%
                    \begin{minipage}{\linewidth}
                        \caption{\centering\footnotesize
                            \mbox{Static} Dis\-tribution
                            \label{fig:prodist:original}}
                    \end{minipage}}
            }
        }
        \noindent\makebox[\linewidth][c]{%
            \raisebox{-70pt}{\rotatebox[origin=c]{45}{%
                    \begin{minipage}{\linewidth}
                        \caption{\centering\footnotesize
                            \mbox{Bayesian} \mbox{Updates}
                            \label{fig:prodist:updates}}
                    \end{minipage}}
            }
        }
    \end{subfigure}%
    \begin{subfigure}[t]{.21\textwidth}
        \centering
        \minibox[c]{\scriptsize 1\textsuperscript{st} sample \vspace{1.5mm}}
        \includegraphics[width=\linewidth]
        {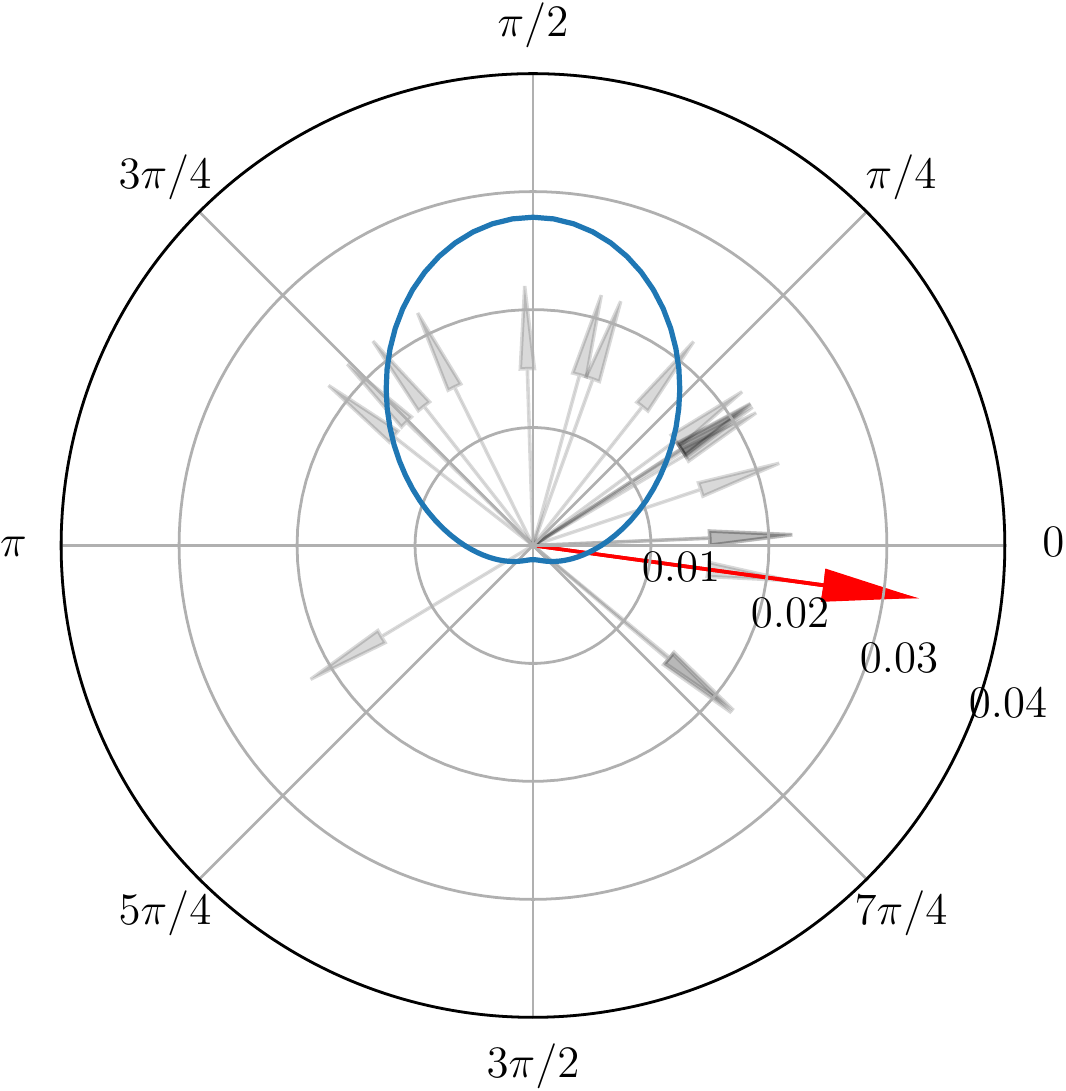}
        \includegraphics[width=\linewidth]
        {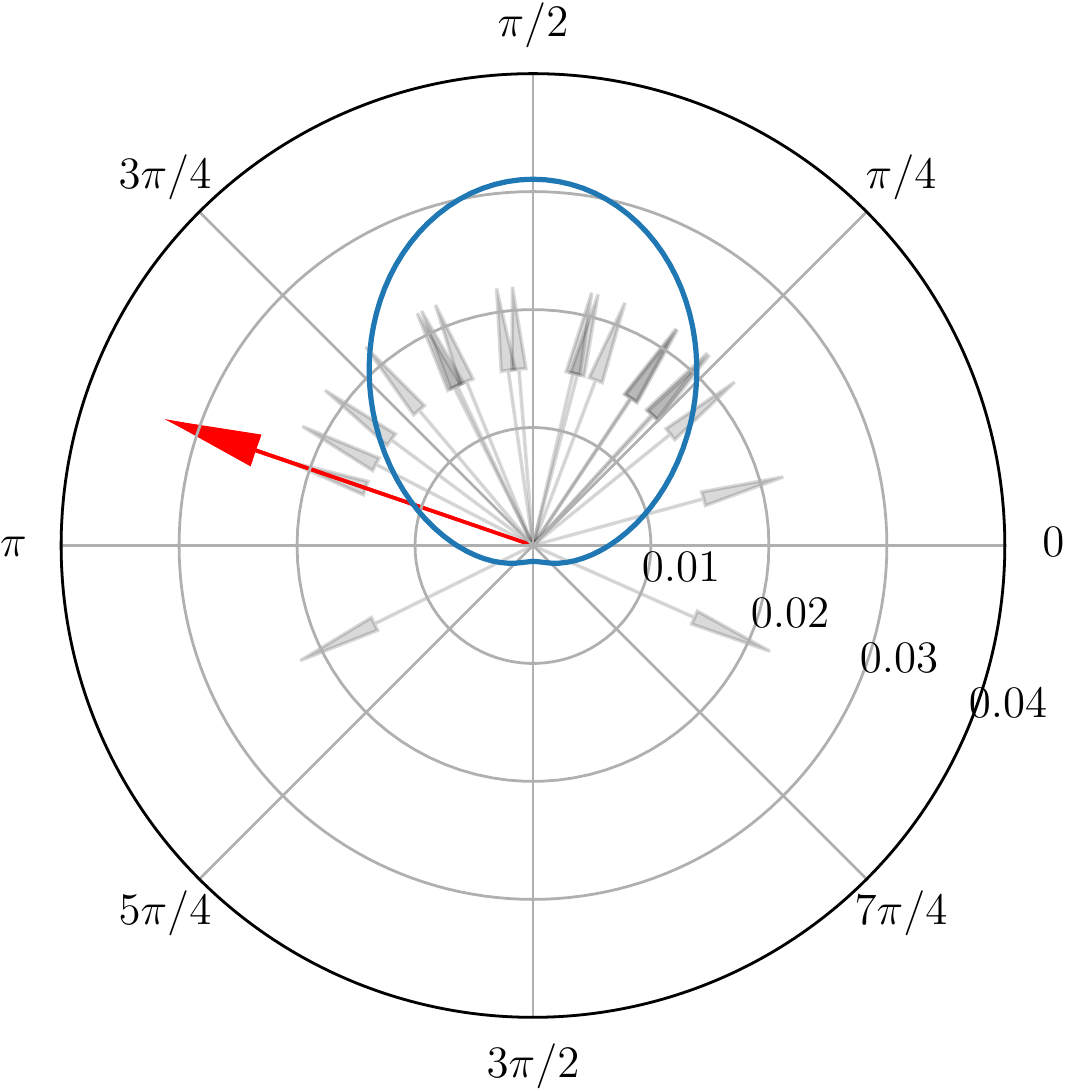}
    \end{subfigure}%
    \begin{subfigure}[t]{.21\textwidth}
        \centering
        \minibox[c]{\scriptsize 5\textsuperscript{th} sample \vspace{1.5mm}}
        \includegraphics[width=\linewidth]
        {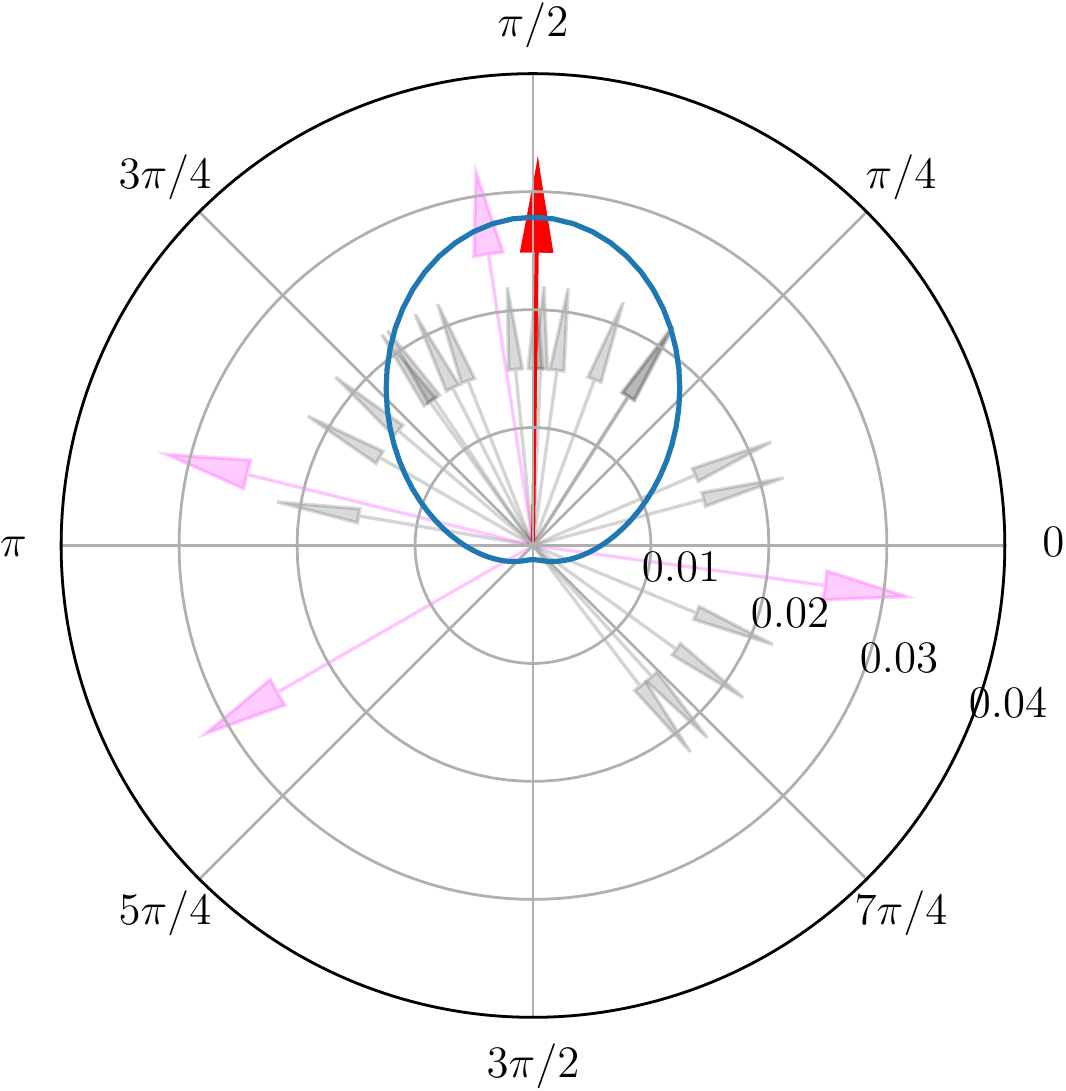}
        \includegraphics[width=\linewidth]
        {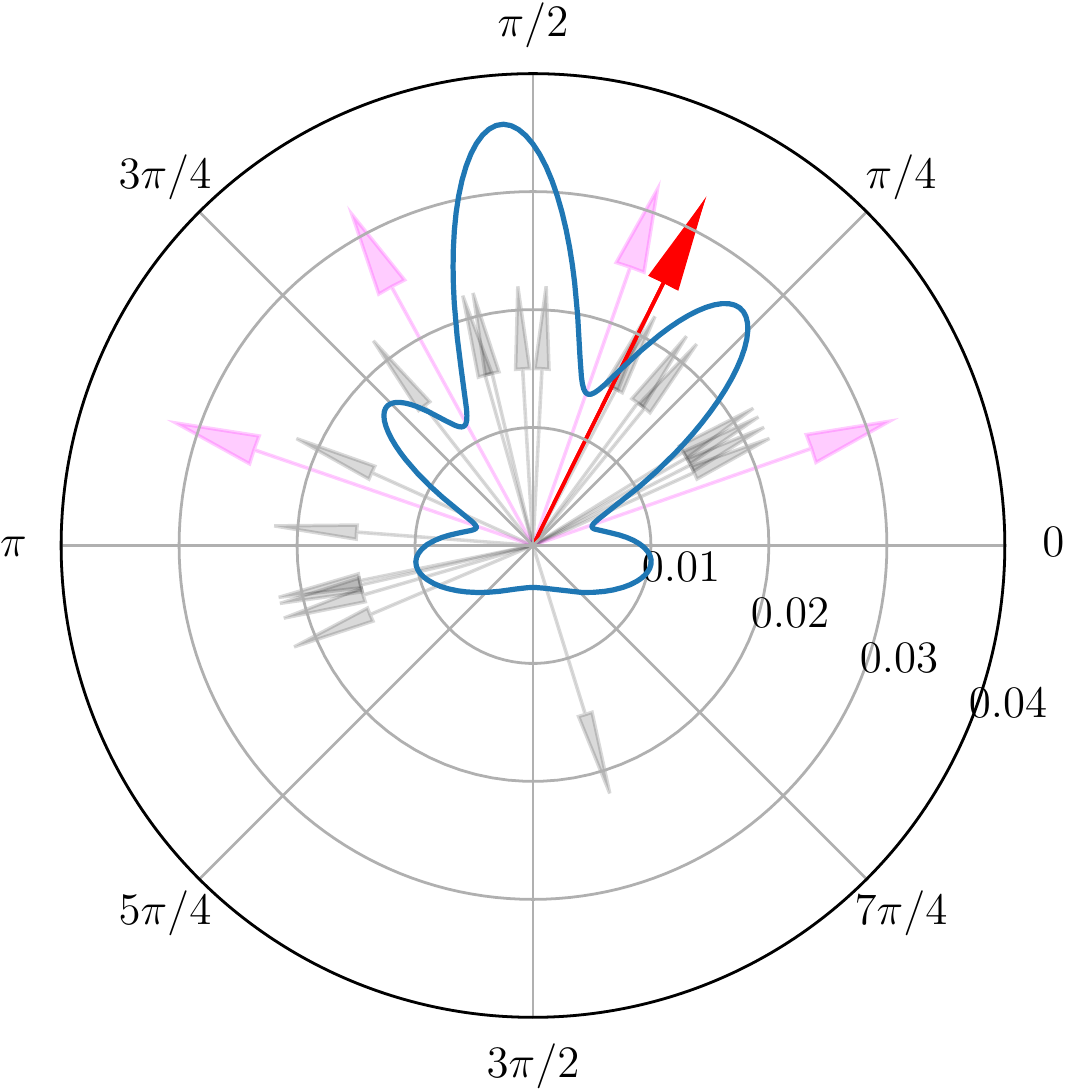}
    \end{subfigure}%
    \begin{subfigure}[t]{.21\textwidth}
        \centering
        \minibox[c]{\scriptsize 10\textsuperscript{th} sample \vspace{1.5mm}}
        \includegraphics[width=\linewidth]
        {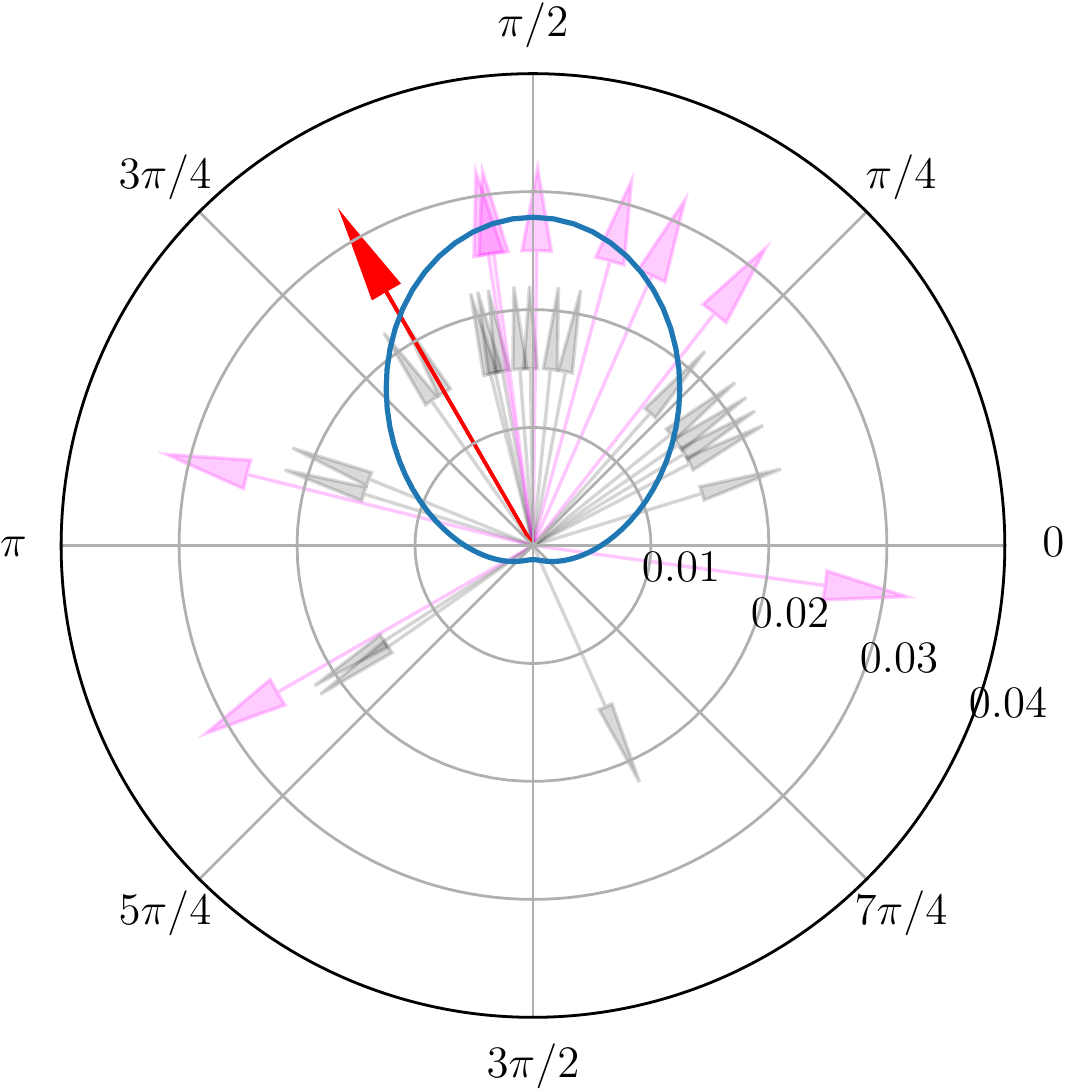}
        \includegraphics[width=\linewidth]
        {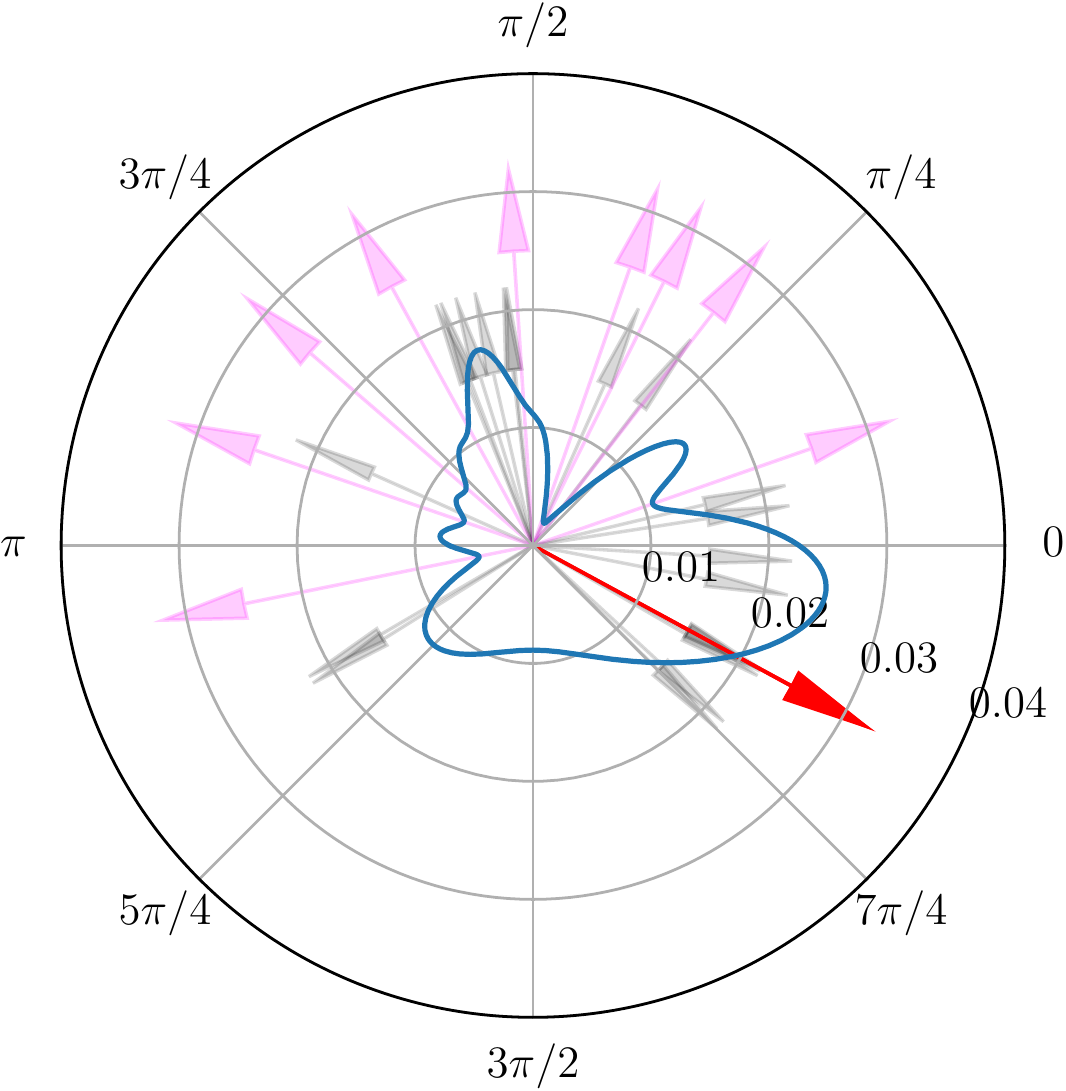}
    \end{subfigure}%
    \begin{subfigure}[t]{.21\textwidth}
        \centering
        \minibox[c]{\scriptsize 15\textsuperscript{th} sample \vspace{1.5mm}}
        \includegraphics[width=\linewidth]
        {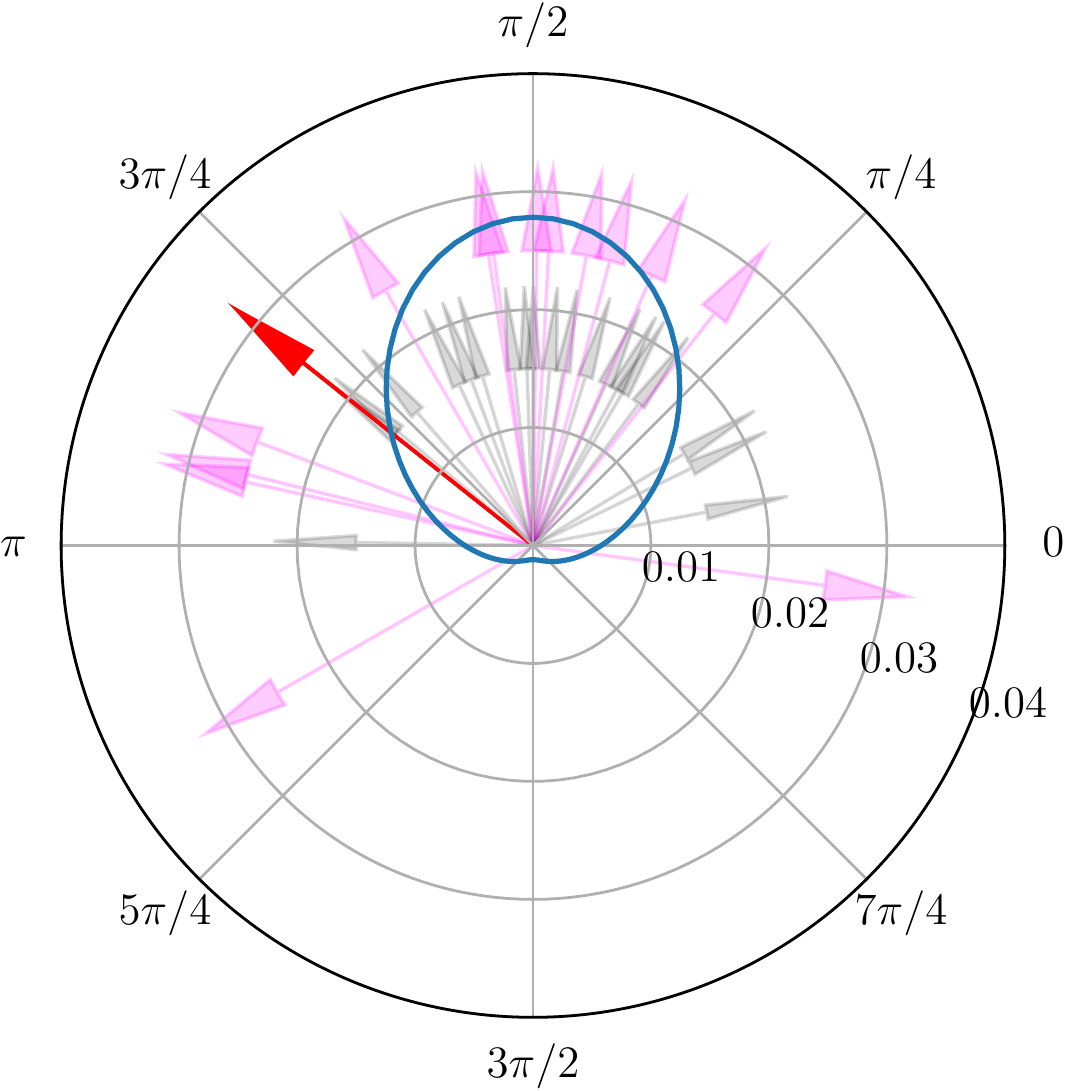}
        \includegraphics[width=\linewidth]
        {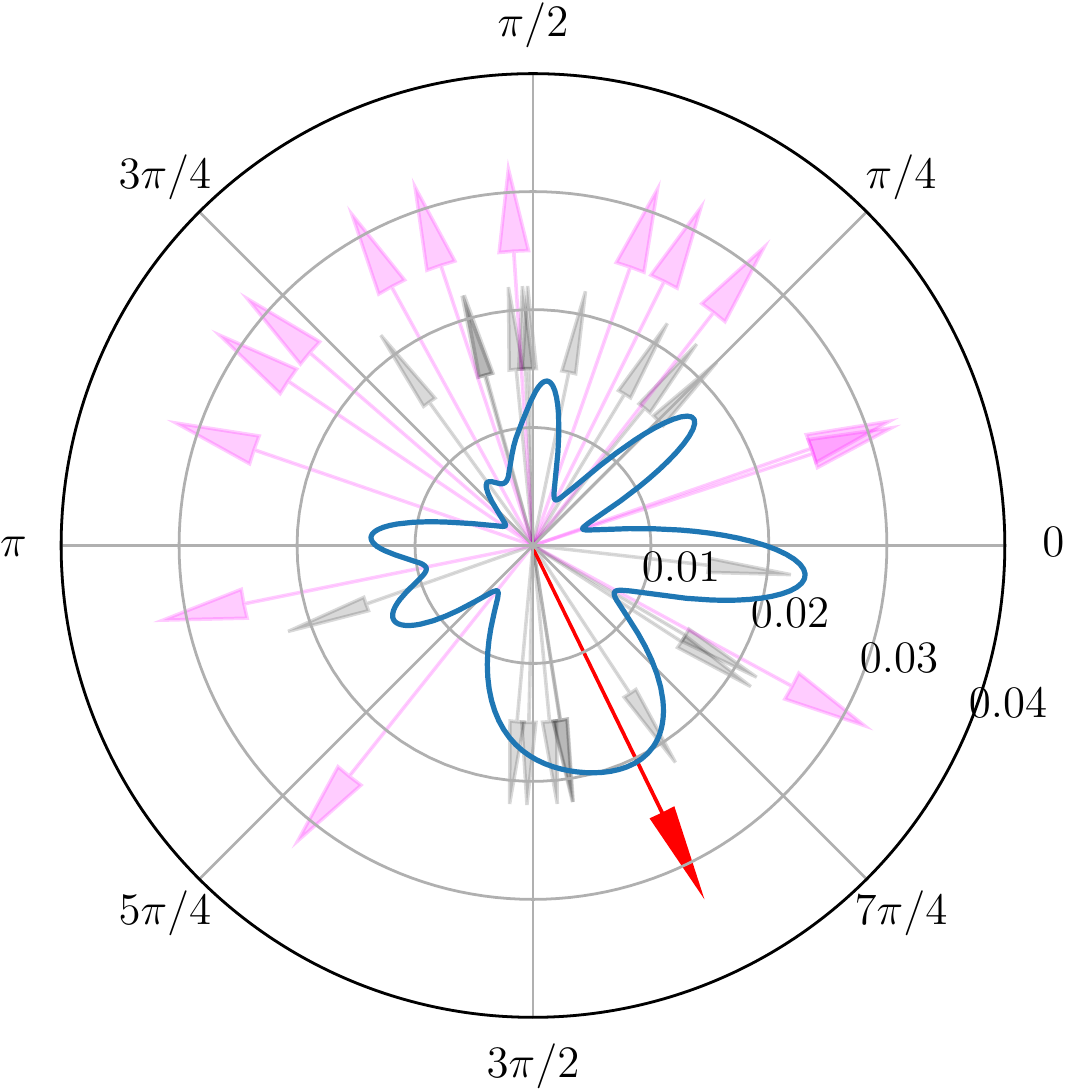}
    \end{subfigure}
    \caption{\footnotesize
        Comparison between a static distribution and our proposing distribution with sequential Bayesian updates, demonstrating what happens when a local planner failed to extends towards the sampled directions every single time.
        Only every 5\textsuperscript{th} time step is shown for conciseness.
        The \textbf{\color{alizarin}red} arrow represents the sampled directions at the given time-step,
        the \textbf{\color{amethyst}purple} arrows represent all sampled (and failed) directions so far, and the \textbf{\color{ashgrey}grey} arrows illustrate the likelihood of sampling 20 times from the distribution shown.
        (\subref{fig:prodist:original})
        Redrawing samples from a static distribution will be probabilistically complete in the limit, however, is wasteful for drawing from the same probability density.
        (\subref{fig:prodist:updates})
        Initially, it behaves very similar to the underlying proposal distribution (von Mises-Fisher), but when a sampled direction is invalid (blocked by obstacles) it utilises a kernel to sequentially update the proposal distribution by incorporating the failure information.
        \label{fig:prodist}
    }
\end{figure*}
}

Any periodic kernel $k(\cdot)$ that has the property of measuring similarity between two $x$ can be applied to~\cref{eq:recursive-def-sample-dist}, as long as $k(x, x') \le 1 \, \forall x' \in \mathbb{R}$.
In particular, we employ the periodic squared exponential kernel~\autocite{mackay1998_IntrGaus} in our settings, which is given by
\begin{equation}
    k_\text{SEP}(x,x') = \sigma^2 \exp\left({-\frac{2\sin^2(\pi(x-x')/p)}{\lambda^2}}\right),
\end{equation}
where $\lambda$ is the length scale, $\sigma$ is a scaling factor, and we set $p=2\pi$ as the period of repetition.
The periodic squared exponential kernel is chosen as its periodic nature is necessary for our usage in spherical distribution.
Alternative kernels, such as the exponential or the Laplacian kernel can also be utilised if they are periodic in the spherical domain.



In practice, since we use this kernel to sequentially decrease the probability density $\SampleDist$ when we observe a failed sample, the parameters $\sigma$ and $\lambda$ characterise the magnitude and the surrounding regions that the Bayesian updates affects from observing $x'$, respectively.
During updates, we scale the magnitude of kernel with $\sigma = \sqrt{\beta}$ where $\beta\in\mathbb{R},\, 0 < \beta \le 1$ which acts as a factor to control the influence of kernel.
Therefore, we can rewrite~\cref{eq:recursive-def-sample-dist} as
\begin{equation}\label{eq:full-proposing-baye-samp-dist}
    \SampleDist_i(x \given x_{i-1},\FailedSamSet_j)
     = \frac{
        \splitfrac{
            \SampleDist_i\Big(x \given[\big] x_{i-1},\FailedSamSet_{j-1}\Big)
            \Big(1 -
        }{
            \beta\exp\Big({-\frac{2\sin^2((x-x'_{j-1})/2)}{\lambda^2}}\Big)
            \Big)
        }
    }{\alpha_{j}} \;\forall\, j > 1
\end{equation}

which acts as our sequential update rule for the $j$\textsuperscript{th} samples in state $\State_i$.

The overall sampling procedure is give in~\Cref{alg:bayesian-local-sampling}.
At each iteration, we first check if any of the arm (local sampler) has low probability (in the MAB settings), of which such an arm will be restarted at a new location to facilitate exploration (\cref{alg:rrdt:restart-arm}).
The procedure is same as Algorithm 1 in~\autocite{lai2018_BalaGlob}, which will results in a new node.
Otherwise, an arm $k\in\mathcal{K}$ is picked with a MAB scheduler, and proceed to perform the proposed Bayesian local-sampling method (line \ref{alg:rrdt:baye-sam-start} to \ref{alg:rrdt:baye-sam-end}).
Instead of using a static distribution in~\autocite{lai2018_BalaGlob}, our method adaptively adjust the distribution.
If $k$ has no previous history (i.e. first initialisation) we draws a new direction uniformly;
otherwise, we perform a chained sampling from $k$ by conditioning on the last immediate successful direction and the set of previous failed directions.
As a result, the sequential Bayesian updates scheme at \cref{alg:rrdt:seq-update} utilises kernel methods to encapsulate the likelihood to draw new samples at a more promising direction.
The $\epsilon_n$ in~\cref{alg:rrdt:step-towards} refers to the vanishing radius $\epsilon_n = \gamma(\frac{\log(n)}{n})^{1/d}$ given in~\autocite{karaman2010_IncrSamp} which ensure optimality as it creates finer connections as the size of the graph increases.

\begin{table}[b]
    \tabcolsep=0.11cm
    \caption{Experimental results.
    Sampled points are in the format $(\mu\pm\sigma)\cdot10^3$.
    Numbers below environments represent nodes budget.
    \label{table:env-description}}
    \vspace{-2mm}
    \resizebox{\linewidth}{!}{%
    \begin{tabular}{@{}ccccccc@{}}
\toprule
Env.        &                               & RRT*                  & Bi-RRT*               & Informed-RRT*         & RRdT*                 & RRdT* (baye.seq.)     \\ \midrule
Room        & \multicolumn{1}{c|}{Samp.Pt.} & $23\pm0.7$  & $22\pm0.7$  & $23\pm0.5$  & $20\pm0.7$  & $20\pm0.2$  \\
(10000)     & \multicolumn{1}{c|}{Succ.}    & 100\%                 & 100\%                 & 100\%                 & 100\%                 & 100\%                 \\ \midrule
Maze        & \multicolumn{1}{c|}{Samp.Pt.} & $304\pm24$  & $248\pm19$  & $296\pm19$  & $89\pm0.4$  & $64\pm0.1$  \\
(50000)     & \multicolumn{1}{c|}{Succ.}    & 40\%                  & 80\%                  & 30\%                  & 100\%                 & 100\%                 \\ \midrule
Clutter     & \multicolumn{1}{c|}{Samp.Pt.} & $628\pm42$ & $532\pm24$ & $596\pm41$  & $124\pm0.6$  & $98\pm0.2$  \\
(50000)     & \multicolumn{1}{c|}{Succ.}    & 15\%                  & 30\%                  & 20\%                  & 90\%                  & 100\%                 \\ \midrule
Manipulator & \multicolumn{1}{c|}{Samp.Pt.} & $540\pm226$ & $379\pm46$  & $635\pm244$ & $164\pm1.6$ & $153\pm1.4$ \\
(20000)     & \multicolumn{1}{c|}{Succ.}    & 0\%                   & 40\%                  & 0\%                   & 100\%                 & 100\%                 \\ \bottomrule
\end{tabular}

    }
\end{table}

\subsection{Drawing Samples from the Proposal Distribution}

Standard SBPs typically employ a global uniform proposal distribution, which ensures probabilistic completeness, but also is cheap to evaluate. Hence, an important aspect in the design of $\mathcal{Q}$ is the computational resources to draw a sample.

The non-trivial probability distribution presented in~\cref{eq:recursive-def-sample-dist}  has no closed-form expression. There exist many methods to sample from $\SampleDist$, for example, adaptive importance sampling or Metropolis-Hastings sampling.
However, these methods are not suitable for sampling from non-stationary distributions and tend to require be more computationally intensive.
The naive way of maintaining a stationary distribution in the original approach \autocite{lai2018_BalaGlob} performs reasonably well without any adaptations and, therefore, if sampling from $\SampleDist$ is highly computationally expensive, then it might not be justifiable.

Therefore, in practice we approximate $\SampleDist$ with a multinomial distribution $\SampleDistMul$, where $\SampleDistMul(\hat{x}) = \Prob(\hat{X}_1=\hat{x}_1, \ldots, \hat{X}_N=\hat{x}_N)$ is a joint distribution on $\hat{X}_1=\hat{x}_1, \ldots, \hat{X}_N=\hat{x}_N$.
The $\SampleDistMul$ resembles a discretised $\SampleDist$ with $\hat{X}_i$ as the random variables that denote the outcomes of the trials by observing results from local-sampling.
Since $\SampleDist$ is a periodic function, its support lies within the interval of $x \in [-\pi, \pi)^{d-1}$.
Let $\SampleDistDelta$ be a sufficiently small positive value where $\SampleDistDelta \ll 2\pi$ and $\SampleDistDelta/2\pi \approx 0$.
We define $\hat{X}_1=\hat{x}_1, \ldots, \hat{X}_N=\hat{x}_N$ to represent the discretised support of $\SampleDist$ in the $(d-1)$-dimensional space, with $\SampleDistDelta$ as the interval between the support.
Then, we say that in the limit
\begin{equation}
    \lim_{\SampleDistDelta \to 0}
    \mathbb{E} [ \SampleDistMul_i(\hat{x} \given \hat{x}_{i-1}, \hat{\FailedSamSet}_j) ]
    \simeq
    \mathbb{E} [ \SampleDist_i(x \given x_{i-1}, \FailedSamSet_j) ]
    ~ \forall\, i,j \ge 1.
\end{equation}
It is trivial to draw a sample $\hat{x}_i \sim \SampleDistMul_i(\hat{x} \given \hat{x}_{i-1}, \hat{\FailedSamSet}_j)$ and, in order to draw a sample continuously in the domain, we mix it with another uniform random variable in the interval of $\SampleDistDelta$ such that $
    x_i = \hat{x_i} + \tilde{x_i}
    \given[\Big]
    \hat{x_i} \sim \SampleDistMul_i(\cdot)
    ,\,
    \tilde{x_i} \sim \mathcal{U}(0, \SampleDistDelta)
$.

\section{Experimental Results}

Our sequential Bayesian distribution, as it observes pasts events, is shown in~\cref{fig:prodist}.
The original approach~\autocite{lai2018_BalaGlob} only conditioned the distribution $h_i(x \given x_{i-1})$ on previous successful direction (\cref{fig:prodist:original}).
Our proposing method performs sequential updates on our distribution $\SampleDist_i(x \given x_{i-1}, \FailedSamSet_j)$ by conditioning on the set of failed sampled directions (\cref{fig:prodist:updates}).
It shows our proposal distribution $\SampleDist$ evolves as more sampled directions that has no free space to extends connections are observed.
The diagrams shown assume the unit vector towards $\frac{\pi}{2}$ is the previous successful sampled direction, with parameters $\beta = 0.9$ and $\lambda = \frac{\pi}{4}$.
The parameter $\beta$ denotes the percentage of probability to decrease at the failed direction.
In general, $\beta$ with a value between $0.8$ and $0.95$ are reasonable to substantially lower the probability of drawing again in the same failed direction.
The parameter $\lambda$ controls how aggressive the kernel is in relation to the nearby region, for example, probability of directions within $\lambda$ radian will decrease at least $62\%$.
In our experiments, we found that $\lambda$ with a value of $\frac{\pi}{8}$ to $\frac{\pi}{2}$ appears to work well in most cases.


    The \RRdT* algorithm with Bayesian updates had been tested empirically against other state-of-the-art sampling-based planners.
    The experimented scenarios are
    (\subref{fig:experiments:room}) \emph{Room}, (\subref{fig:experiments:maze}) \emph{Maze}, (\subref{fig:experiments:clutter}) \emph{Clutter}, and
    (\subref{fig:experiments:manipulator}) \emph{Manipulator}, as shown in~\cref{fig:all-expriment-envs,fig:robotic-arm-expr}.
    The first three scenarios are 2d path finding problems with increasing complexity, and the \emph{manipulator} is a 6 dof scenario of a TX90 robot arm with an attached PR2 gripper.
    
    The metrics obtained after repeating 20 times is shown in \cref{fig:experiments}, and \cref{table:env-description} listed the success rate and the total sampled configurations.
    \emph{Invalid connections} refers to scenarios when the tree tries to extends toward some free space but collides with obstacles in between.
    \emph{Invalid local sampling} refers to local planners proposed a direction that does not result in an extension, hence does not apply to SBPs other than \RRdT* as others do not perform local planning.
    The \emph{invalid metrics} show the proposed algorithm requires fewer samples to achieve the same result as other algorithms, and by invalid it refers to the sampled points that do not result in additional nodes or edges to the graph.
    All plots in~\cref{fig:experiments} are the lesser the better.

    Agreeing with results obtained from the original work, \RRdT outperforms other SBPs as the complexity of the space increases.
    Furthermore, \RRdT* that utilises Bayesian learning of the proposal distribution outperforms the original (stationary proposal) \RRdT* in all of the tested scenarios, as our proposal distribution takes full advantage of information obtained during sampling.
    The effectiveness of modelling Bayesian proposal distribution seems to have decreased in the high dof scenario, as it takes more sample points to model a better distribution.
    However, the sequential updating of our Bayesian distribution does not seem to contribute to any noticeable computational overhead (see supplement).
    Moreover, the proposed method converges to the optimal value as the number of node increases, but tends to return a solution faster in complex scenarios.

\section{Conclusion}

    In conclusion, we present a sequential Bayesian proposal distribution that improves the sample-efficiency for the local sampling-based planning problem.
    The formulation follows directly from the nature of the problem as a Markovian process, in which we take a sequential approach to learn from past events.
    The use of the proposal distribution requires almost no computational overheads, while still improving the sampling success rate.
    Being sample-efficient is essential in robot applications where evaluating the validity of a given sample might be expensive; hence, the reduction in invalid samples, without noticeable overhead, improves our current approach in performing robot planning.

\printbibliography

\end{document}